\documentclass{article}

\usepackage[final,nonatbib]{neurips_2022}

\usepackage[utf8]{inputenc} 
\usepackage[T1]{fontenc}  
\usepackage{xcolor}     
\usepackage{color}
\definecolor{ggreen}{HTML}{58a55c}
\usepackage[pagebackref=true,
   breaklinks=true,
   letterpaper=true,
   colorlinks = true,
   linkcolor = red,
   urlcolor = blue,
   citecolor = ggreen,
   anchorcolor = blue]{hyperref}
\usepackage{url}      
\usepackage{booktabs}    
\usepackage{amsfonts}    
\usepackage{nicefrac}    
\usepackage{microtype}   

\usepackage{amsmath}
\usepackage{graphicx}
\usepackage{algorithm}
\usepackage{subcaption}
\usepackage{xcolor}
\usepackage{colortbl}

\usepackage{pifont}
\usepackage{tabu}
\usepackage{tabularx}
\definecolor{Gray}{gray}{0.9}
\definecolor{Cyan}{rgb}{0.88,1,1}
\usepackage{listings}
\usepackage{bm}
\usepackage{adjustbox}
\usepackage{ragged2e}
\usepackage{multirow}
\usepackage{amssymb}
\usepackage{wrapfig}
\usepackage{makecell}

\usepackage{soul}
\sethlcolor{Gray}

\numberwithin{equation}{section}

\newcommand{\paragrapha}[2][1pt]{\vspace{#1}\noindent\textbf{#2}}

\newcolumntype{x}[1]{>{\centering\arraybackslash}p{#1pt}}
\newlength\savewidth

\newcommand{\PreserveBackslash}[1]{\let\temp=\\#1\let\\=\temp}
\newcolumntype{C}[1]{>{\PreserveBackslash\centering}p{#1}}
\newcolumntype{L}[1]{>{\PreserveBackslash\raggedright}p{#1}}

\usepackage{etoolbox}
\makeatletter
\AfterEndEnvironment{algorithm}{\let\@algcomment\relax}
\AtEndEnvironment{algorithm}{\kern2pt\hrule\relax\vskip3pt\@algcomment}
\let\@algcomment\relax
\newcommand\algcomment[1]{\def\@algcomment{\footnotesize#1}}
\renewcommand\fs@ruled{\def\@fs@cfont{\bfseries}\let\@fs@capt\floatc@ruled
 \def\@fs@pre{\hrule height.8pt depth0pt \kern2pt}%
 \def\@fs@post{}%
 \def\@fs@mid{\kern2pt\hrule\kern2pt}%
 \let\@fs@iftopcapt\iftrue}
\makeatother

\newcommand*\samethanks[1][\value{footnote}]{\footnotemark[#1]}

\newcommand\cb[1]{\color{blue} #1}

\title{HorNet: Efficient High-Order Spatial Interactions with
Recursive Gated Convolutions}

\author{
Yongming Rao$^{1}$\thanks{Equal contribution. ~~\textsuperscript{\dag}Corresponding authors.}
~~
Wenliang Zhao$^{1}$\samethanks
~~~
Yansong Tang$^{1}$ 
\\
\textbf{Jie Zhou}$^{1\dagger}$
~~
\textbf{Ser-Nam Lim}$^{2\dagger}$
~~
\textbf{Jiwen Lu}$^{1\dagger}$
\\ 
$^{1}$Tsinghua University ~~ $^{2}$Meta AI
}

\newcommand{\gnconv}{\ensuremath{\textit{g}^\textit{n}\text{Conv}}}
\newcommand{\gconv}{$\textit{g}\text{Conv}$}
\DeclareMathOperator{\flops}{FLOPs}
\DeclareMathOperator{\IE}{IE}

\begin{document}

\maketitle

\begin{abstract}
 Recent progress in vision Transformers exhibits great success in various tasks driven by the new spatial modeling mechanism based on dot-product self-attention. In this paper, we show that the key ingredients behind the vision Transformers, namely input-adaptive, long-range and high-order spatial interactions, can also be efficiently implemented with a convolution-based framework. We present the Recursive Gated Convolution ($\textit{g}^\textit{n}$Conv) that performs high-order spatial interactions with gated convolutions and recursive designs. The new operation is highly flexible and customizable, which is compatible with various variants of convolution and extends the two-order interactions in self-attention to arbitrary orders without introducing significant extra computation. $\textit{g}^\textit{n}$Conv can serve as a plug-and-play module to improve various vision Transformers and convolution-based models. 
 Based on the operation, we construct a new family of generic vision backbones named HorNet. Extensive experiments on ImageNet classification, COCO object detection and ADE20K semantic segmentation show HorNet outperform Swin Transformers and ConvNeXt by a significant margin with similar overall architecture and training configurations. HorNet also shows favorable scalability to more training data and larger model sizes. Apart from the effectiveness in visual encoders, we also show $\textit{g}^\textit{n}$Conv can be applied to task-specific decoders and consistently improve dense prediction performance with less computation. Our results demonstrate that $\textit{g}^\textit{n}$Conv can be a new basic module for visual modeling that effectively combines the merits of both vision Transformers and CNNs. Code is available at \url{https://github.com/raoyongming/HorNet}.
\end{abstract}

\section{Introduction}

 Convolutional neural networks (CNN) have driven remarkable progress in deep learning and computation vision since the introduction of AlexNet~\cite{krizhevsky2012alex} in the last decade. There are quite a few nice properties of CNNs making them naturally suitable for a wide range of vision applications. Translation equivariance introduces useful inductive biases to major vision tasks and enables transferability across different input resolutions. The highly optimized implementation makes it efficient on both high-performance GPUs and edge devices. The evolution of architectures~\cite{lenet,krizhevsky2012alex,simonyan2014very,szegedy2015going,he2016deep,howard2017mobilenets,tan2019efficientnet} further increases its popularity on various vision tasks. 
 
 The emergence of Transformer-based architectures~\cite{dosovitskiy2020vit,touvron2020deit,liu2021swin} greatly challenges the dominance of CNNs. By combining some successful designs in CNN architectures and the new self-attention mechanism, vision Transformers have shown leading performance on various vision tasks such as image classification~\cite{dai2021coatnet,liu2021swin,riquelme2021scaling}, object detection~\cite{zhang2022dino,liu2021swinv2}, semantic segmentation~\cite{cheng2021masked,cheng2021maskformer} and video understanding~\cite{yan2022multiview,fan2021multiscale}. \emph{What makes vision Transformers more powerful than CNNs?} Some efforts have been made to improve the CNN architectures by learning from the new designs in vision Transformers. \cite{liu2022convnet} presents a thorough study to adopt the \emph{meta architecture} of vision Transformer to improve CNNs and proposes to use a \emph{large 7$\times$7 kernel} to construct a modern CNN. \cite{rao2021global} and \cite{ding2022scaling} propose to use even \emph{larger kernels} to learn long-range relations with global filters and up to 31$\times$31 convolutions, respectively. \cite{han2021demystifying} shows that the \emph{input-adaptive weights} play a key role in vision Transformers and achieve similar performance with Swin Transformers with dynamic convolutions~\cite{chen2020dynamic,jia2016dynamic}. However, the effectiveness of dot-product self-attention in vision tasks has not been analyzed from the prospective of \emph{high-order spatial interactions}. 
 
\begin{figure}[t]
  \centering
  \includegraphics[width=\textwidth]{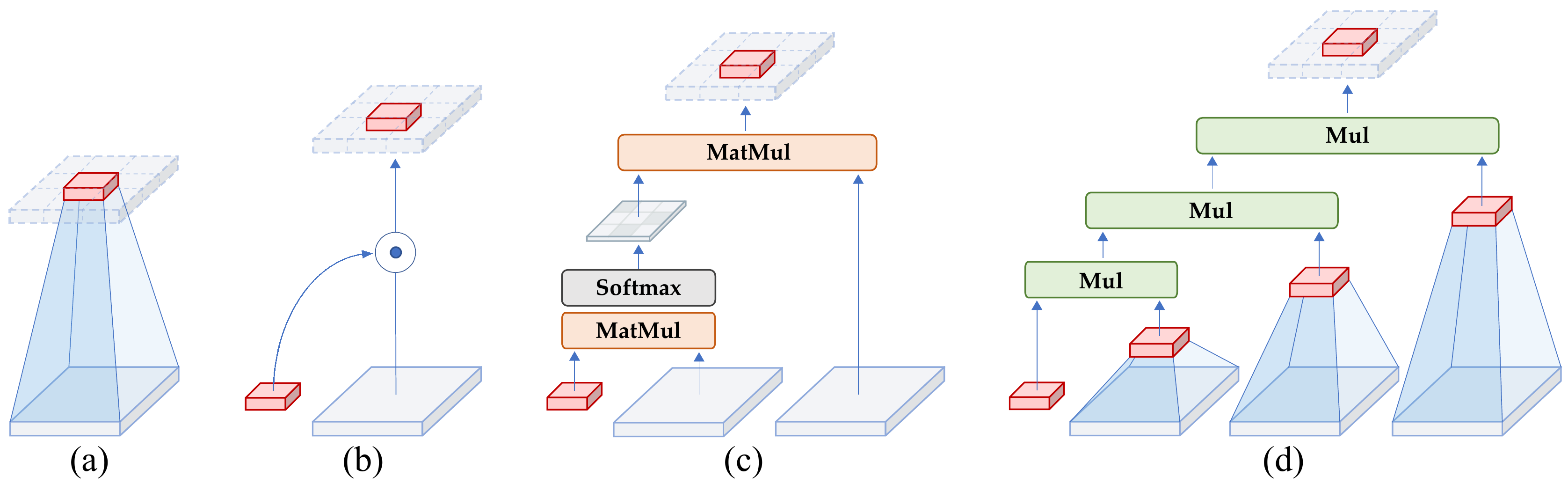}
  \caption{\small \textbf{Illustration of our main idea.} We show representative spatial modeling operations that perform different orders of interactions. In this paper, we focus on studying \emph{explicit} spatial interactions between a feature (red) and its neighboring region (light gray). (a) The standard convolution operation does not explicitly consider the spatial interaction. (b) Dynamic convolution~\cite{jia2016dynamic,chen2020dynamic} and SE~\cite{senet} introduce the dynamic weights to improve the modeling power of convolutions with extra spatial interactions. (c) The self-attention operation~\cite{Vaswani2017transformer} performs two-order spatial interactions with two successive matrix multiplications. (d) $\textit{g}^\textit{n}$Conv realizes arbitrary-order spatial interactions using a highly efficient implementation with gated convolutions and recursive deigns. }
  \label{fig:idea} \vspace{-15pt}
\end{figure}

 While there exists complex and often high-order interactions between two spatial locations in a deep model due to the non-linearity, the success of self-attention and other dynamic networks suggests that the \emph{explicit} and \emph{high-order} spatial interactions introduced by the architectural designs are beneficial to improving the modeling power of vision models. As illustrated in Figure~\ref{fig:idea}, the plain convolution operation does not explicitly consider the spatial interactions between a spatial location (\ie, the red feature) and its neighboring region (\ie, the light gray region). Enhanced convolution operations like dynamic convolution~\cite{chen2020dynamic,jia2016dynamic,han2021demystifying} introduce explicit spatial interaction by generating dynamic weights. The dot-product self-attention operation in Transformers~\cite{Vaswani2017transformer} consists of two successive spatial interactions by performing matrix multiplication among queries, keys and values. The trend of the basic operations for visual modeling indicates that the network capacity can be improved by increasing the order of spatial interactions. 
 
 In this paper, we summarize that the key ingredient behind the success of vision Transformers is the new way of spatial modeling with \emph{input-adaptive}, \emph{long-range} and \emph{high-order} spatial interactions performed by the self-attention operation. While previous work has successfully migrated the meta architecture~\cite{liu2022convnet,han2021demystifying,rao2021global,ding2022scaling}, input-adaptive weight generation strategy~\cite{han2021demystifying} and large-range modeling ability~\cite{rao2021global,ding2022scaling} of vision Transformers to CNN models, a higher-order spatial interaction mechanism has not been studied. We show that all the three key ingredients can be efficiently implemented using a convolution-based framework. We propose the Recursive Gated Convolution ($\textit{g}^\textit{n}$Conv) that performs high-order spatial interactions with gated convolutions and recursive deigns. Instead of simply imitating the successful designs in self-attention, $\textit{g}^\textit{n}$Conv has several extra favorable properties: 1) \emph{Efficient}. The convolution-based implementation avoids the quadratic complexity of self-attention. The design that progressively increases the channel width during performing spatial interactions also enables us to achieve higher-order interactions with bounded complexity; 2) \emph{Extendable}. We extend the two-order interaction in self-attention to arbitrary orders to further improve the modeling power. Since we do not make assumptions on the type of spatial convolution, $\textit{g}^\textit{n}$Conv is compatible with various kernel size and spatial mixing strategies like~\cite{rao2021global,ding2022scaling}; 3) \emph{Translation-equivariant}. $\textit{g}^\textit{n}$Conv fully inherits the translation equivariance of the standard convolution, which introduces beneficial inductive biases to major vision tasks and avoids the asymmetry brought by local attention~\cite{liu2021swin,li2022exploring}. 
 
 Based on $\textit{g}^\textit{n}$Conv, we construct a new family of generic vision backbones named HorNet. We conduct extensive experiments on ImageNet classification~\cite{deng2009imagenet}, COCO object detection~\cite{lin2014coco} and ADE20K semantic segmentation~\cite{zhou2017scene} to verify the effectiveness of our models. With the same 7$\times$7 kernel/window and similar overall architecture and training configurations, HorNet outperforms Swin and ConvNeXt by a large margin on all tasks at different levels of complexity. The gap can be further enlarged by using a global kernel size~\cite{rao2021global}. HorNet also shows favorable scalability to more training data and larger model size, attaining 87.7\% top-1 accuracy on ImageNet, 57.9\% mIoU on ADE20K val and 59.2\% bounding box AP on COCO val with ImageNet-22K pre-training. Apart from applying \gnconv{} in visual encoders, we further test the generality of our designs on task-specific decoders. By adding \gconv{} to the widely used feature fusion model FPN~\cite{lin2017fpn}, we develop HorFPN to model the high-order spatial relationships of features from different hierarchical levels. We observe that HorFPN can also consistently improve various dense prediction models with lower computational costs. Our results demonstrate that $\textit{g}^\textit{n}$Conv can be a promising alternative to self-attention for visual modeling and effectively combine the merits of both vision Transformers and CNNs.

\section{Related Work}
\paragrapha{Vision Transformers. } The Transformer architecture~\cite{Vaswani2017transformer} is originally designed for the natural language processing tasks. 
Since Dosovitskiy \etal~\cite{dosovitskiy2020vit} show that vision models constructed only by the Transformer blocks and a patch embedding layer can also achieve competitive performance to CNNs, many new models have been proposed to modify the Transformer-based architecture and make it more suitable for various vision tasks~\cite{liu2021swin,wang2021pyramid,wu2021cvt,chu2021twins,yang2021focal,tu2022maxvit}. Different from the original designs in~\cite{dosovitskiy2020vit}, state-of-the-art vision Transformers usually utilize a CNN-like hierarchical architecture and change the global self-attention among all patches to local self-attention to avoid the quadratic complexity. In this paper, we follow the overall architecture of the previous hierarchical vision Transformers~\cite{liu2021swin} and replace the self-attention sub-layer with our proposed $\textit{g}^\textit{n}$Conv to fairly compare with the previous Transformer-based models. 

\paragrapha{Convolution-based models. } Inspired by the recent success of vision Transformers, several papers propose to adopt the Transformer-style architecture and spatial convolutions with a large kernel size to improve the performance of CNNs. Han \etal~\cite{han2021demystifying} replace the window self-attention in Swin Transformers with large-kernel dynamic convolutions and achieve better performance. GFNet~\cite{rao2021global} proposes to perform the global spatial interactions like vision Transformers with global filters in the frequency domain, which are equivalent to depth-wise convolutions with a global kernel size and circular padding. ConvNeXt~\cite{liu2022convnet} thoroughly analyzes the designs in recent vision Transformers and presents a strong convolutional model with 7$\times$7 depth-wise convolutions. RepLKNet~\cite{ding2022scaling} explores CNN models with very large kernels (up to 31$\times$31), showing good scalability as vision Transformers. VAN~\cite{guo2022van} and FocalNet~\cite{yang2022focal} use gated convolutions to perform input-adaptive attention and adopts large-kernel dilated convolutions and multiple successive 3$\times$3 convolutions respectively to produce the weights. Previous work focuses on the meta architecture~\cite{yu2021metaformer}, large-kernel designs and input-adaptive weights to improve CNNs by learning from vision Transformers. In this paper, we offer a new perspective of high-order spatial attention to analyze the merits of vision Transformers. We show that the proposed HorNet that combines the advantages of both CNNs and vision Transformers is a better architecture for various vision tasks.

\paragrapha{Hybrid models. } Combining vision Transformers and CNNs to develop hybrid architectures is a new direction in various visual recognition problems. Recently, several efforts have been made to integrate the two types of blocks into a unified model with a sequential~\cite{dai2021coatnet,jiang2021token,yuan2021t2t,xiao2021early} or parallel~\cite{acmix,cui2022mixformer} design. Many enhanced vision Transformers also use lightweight convolutions in the basic building block to efficiently capture neighboring patterns~\cite{dong2021cswin,wu2021cvt,d2021convit} or relax the quadratic complexity of self-attention~\cite{chu2021twins,wang2021pyramid,fan2021multiscale}. Different from these hybrid models, we aim to develop a self-attention free model while combining the favorable properties of both vision Transformers and CNNs.

\section{Method}
\subsection{$\textit{g}^\textit{n}$Conv: Recursive Gated Convolutions}\label{sec:gconv} 
In this section, we will present \gnconv{}, an efficient operation to achieve long-term and high-order spatial interactions. The \gnconv{} is built with standard convolutions, linear projections and element-wise multiplications, but has a similar function of input-adaptive spatial mixing to self-attention.

\paragrapha{Input-adaptive interactions with gated convolution. } Recent success in vision Transformers mainly depends on the proper modeling of the spatial interactions in visual data. Unlike CNNs that simply use the static convolution kernel to aggregate neighboring features, vision Transformers apply multi-head self-attention to dynamically generate the weights to mix spatial tokens. However, the quadratic complexity w.r.t. the input size of the self-attention largely hinders the application of vision Transformers, especially on downstream tasks including segmentation and detection where higher-resolution feature maps are required. In this work, instead of reducing the complexity of self-attention like previous methods~\cite{liu2021swin,chu2021twins, wang2020linformer}, we seek a more efficient and effective way to perform spatial interactions with simple operations like convolution and fully-connected layers. 

The basic operation of our method is the gated convolution (\gconv). Let $\mathbf{x}\in\mathbb{R}^{HW\times C}$ be the input feature, the output of the gated convolution $\mathbf{y}=\gconv(\mathbf{x})$ can be written as:
\begin{equation}
\begin{split}
   [\mathbf{p}_0^{HW\times C}, \mathbf{q}_0^{HW\times C}]=\phi_{\rm in}(\mathbf{x})\in\mathbb{R}^{HW\times 2C},\\
   \mathbf{p}_1 =f(\mathbf{q}_0)\odot\mathbf{p}_0 \in\mathbb{R}^{HW\times C},\quad 
   \mathbf{y} = \phi_{\rm out}(\mathbf{p}_1)\in \mathbb{R}^{HW\times C},
   \label{equ:gconv} 
\end{split}
\end{equation}
where $\phi_{\rm in},\phi_{\rm out}$ are linear projection layers to perform channel mixing and $f$ is a depth-wise convolution. Note that $p_1^{(i, c)}=\sum_{j\in\Omega_i} w_{i\to j}^c q_0^{(j,c)}p_0^{(i, c)}$, where $\Omega_i$ is the local window centered at $i$ and $w$ represents the convolution weight of $f$. Therefore, the above formulation explicitly introduce interactions among the neighboring features $\mathbf{p}_0^{(i)}$ and $\mathbf{q}_0^{(j)}$ through the element-wise multiplication. We consider the interaction in \gconv{} as \textit{1-order interaction} as each $\mathbf{p}_0^{(i)}$ has interacted with its neighbor feature $\mathbf{q}_0^{(j)}$ only once.

\paragrapha{High-order interactions with recursive gating. } After achieving an efficient 1-order spatial interactions with the \gconv, we then design the \gnconv{}, a recursive gated convolution to further enhance the model capacity by introducing higher-order interactions. Formally, we first use $\phi_{\rm in}$ to obtain a set of projected features $\mathbf{p}_0$ and $\{\mathbf{q}_k\}_{k=0}^{n-1}$:
\begin{equation}
  \left[\mathbf{p}_0^{HW\times C_0},\mathbf{q}_0^{HW\times C_0}, \ldots, \mathbf{q}_{n-1}^{HW\times C_{n-1}}\right]=\phi_{\rm in}(\mathbf{x})\in \mathbb{R}^{HW\times (C_0 + \sum_{0\le k\le n-1} C_k)}.
\end{equation}
We then perform the gated convolution \emph{recursively} by
\begin{equation}
  \mathbf{p}_{k+1}=f_{k}(\mathbf{q}_{k})\odot g_{k}(\mathbf{p}_{k})/\alpha, \qquad k=0, 1, \ldots, n-1, \label{equ:recursive}
\end{equation}
where we scale the output by $1/\alpha$ to stabilize the training. $\{f_k\}$ are a set of depth-wise convolution layers and $\{g_k\}$ are used to match the dimension in different orders:
\begin{equation}
g_k=
  \begin{cases}
  \text{Identity}, &k=0,\\
  \text{Linear}\left(C_{k-1}, C_{k}\right), &1\le k\le n-1.
  \end{cases}
\end{equation}
Finally, we feed the output of the last recursion step $\mathbf{q}_{n}$ to the projection layer $\phi_{\rm out}$ to obtain the result of the \gnconv{}. From the recursive formula Equation~\eqref{equ:recursive}, it is easy to show that the interaction-order of $\mathbf{p}_{k}$ will be increased by 1 after each step. As a result, we can see that the \gnconv{} achieves $n$-order spatial interactions. It is also worth noting that we need only a single $f$ to perform depth-wise convolution to the concatenation of the features $\{\mathbf{q}_k\}_{k=0}^{n-1}$ together instead of computing the convolution in each recursive step as in Equation~\eqref{equ:recursive}, which can further simplify the implementation and improve the efficiency on GPUs. To ensure that the high-order interactions do not introduce too much computational overhead, we set the channel dimension in each order as:
\begin{equation}
  C_k = \frac{C}{2^{n-k-1}}, \qquad 0\le k\le n-1.
\end{equation}
This design indicates that we perform the interactions in a coarse-to-fine manner, where lower orders are computed with fewer channels. Besides, the channel dimension of $\phi_{\rm in}(\mathbf{x})$ is exactly $2C$ and the total FLOPs can be strictly bounded even with $n$ increasing. It can be proved that (see Appendix~\ref{appendix:flops}):
\begin{equation}
  \flops(\gnconv) < HWC(2K^2 + 11/3\times C + 2),\label{equ:upper_bound}
\end{equation}
where $K$ is the kernel size of the depth-wise convolution. Therefore, our \gnconv{} achieves high-order interactions with a similar computational cost to a convolutional layer.

\paragrapha{Long-term interactions with large kernel convolutions. } Another difference between vision Transformers and conventional CNNs is the receptive field. Conventional CNNs~\cite{simonyan2014very,he2016deep} often use 3$\times$3 convolution through the whole network, while vision Transformers calculate self-attention on the whole feature maps~\cite{dosovitskiy2020vit,touvron2020deit} or inside a relatively large local window (\eg, 7$\times$7). The large receptive field in vision Transformers makes it easier to capture long-term dependencies, which is also recognized as one of the key advantages of  vision Transformers. Inspired by this design, there are some efforts to introduce large kernel convolutions to CNNs recently~\cite{ding2022scaling,liu2022convnet,rao2021global}. To make our \gnconv{} capable of capturing long-term interactions, we adopt two implementations for the depth-wise convolution $f$:
\begin{itemize}
  \item \emph{7$\times$7 Convolution}. 7$\times$7 is the default window/kernel size of Swin Transformers~\cite{liu2021swin} and ConvNext~\cite{liu2022convnet}. Studies in~\cite{liu2022convnet} show that the kernel size produces good performance on ImageNet classification and various downstream tasks. We follow this configuration to fairly compare with representative work of vision Transformers and modern CNNs. 
  \item \emph{Global Filter (GF)}. The GF layer~\cite{rao2021global} multiplies the frequency domain features with learnable global filters, which is equivalent to a convolution in the spatial domain with a global kernel size and circular padding. We use a modified version of the GF layer by processing half of the channels with the global filter and the other half with 3$\times$3 depth-wise convolutions and only use GF layers in late stages to preserve more local details. 
\end{itemize}

 \begin{figure}[t]
  \centering
  \includegraphics[width=\textwidth]{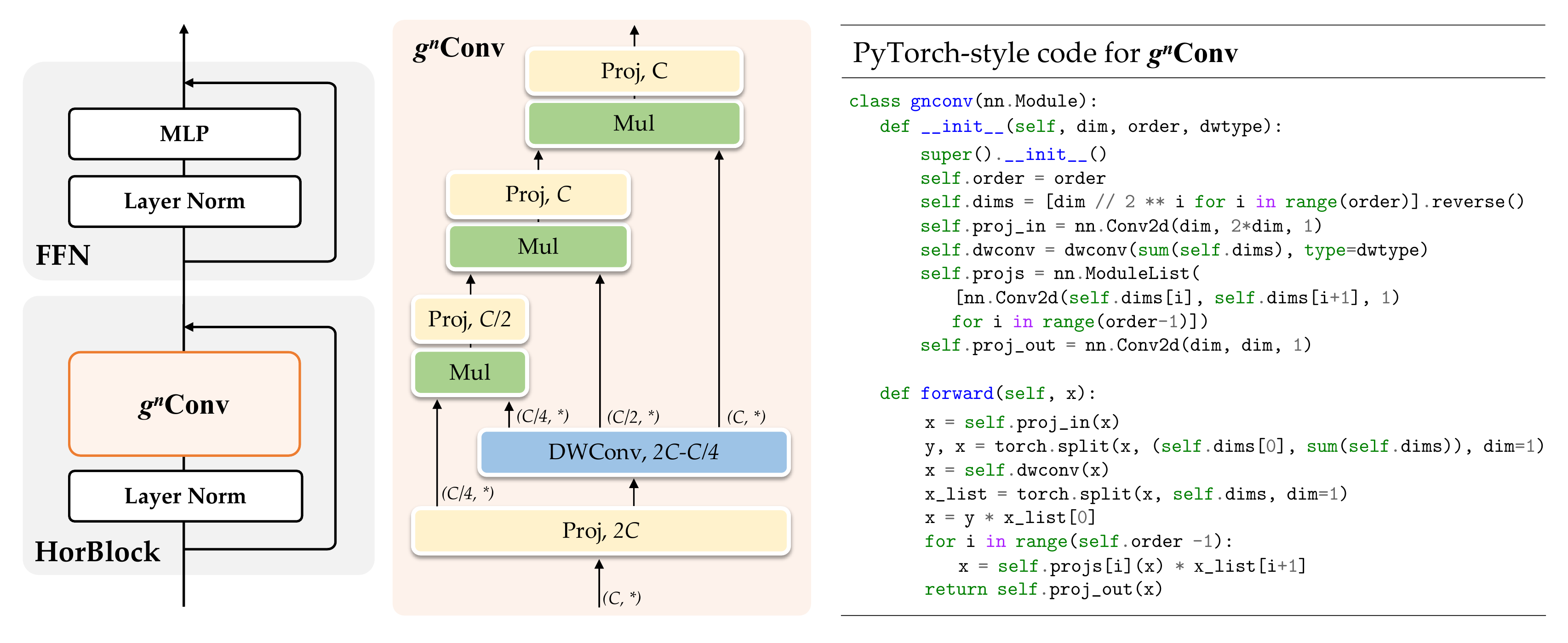}
  \caption{\small \textbf{Overview of the basic building block in HorNet with $\textit{g}^\textit{n}$Conv. } We adopt the block design of Transformers~\cite{Vaswani2017transformer} and replace the self-attention sub-layer with $\textit{g}^\textit{n}$Conv to develop our HorNet (\emph{left}). We also provide the detailed implementation of $\textit{g}^\text{3}$Conv (\emph{middle}) and the Pytorch-style code for an arbitrary order (\emph{right}). }
  \label{fig:block} \vspace{-10pt}
\end{figure}


\paragrapha{Spatial interactions in vision models.} We review some representative vision model designs from the perspective of spatial interactions, as shown in Figure~\ref{fig:idea}. Specifically, we are interested in the interactions between a feature $\mathbf{x}_i$ and its neighboring feature $\mathbf{x}_j, j\in\Omega_i$. By using the tool designed for explaining the interaction effect (IE) in \cite{lerman2021explaining,ai2003interaction}, we provide an intuitive analysis of the order of explicit spatial interactions in Appendix~\ref{appendix:highorder}. Our analysis reveals a key difference between vision Transformers and previous architectures from a new view, \ie,  vision Transformers have higher-order spatial interactions in each basic block. The result inspires us to explore an architecture that can realize more efficient and effective spatial interactions with more than two orders. As discussed above, our proposed \gnconv{} can achieve arbitrary-order interactions with bounded complexity. It is also worth noting that similar to other scaling factors in deep models like width~\cite{zagoruyko2016wide} and depth~\cite{he2016deep}, simply increasing the order of spatial interactions without considering the overall model capacity will not lead to a good trade-off~\cite{tan2019efficientnet}. In this paper, we focus on developing a stronger visual modeling architecture based on the analysis of the spatial interaction orders of well-designed models. We believe a more thorough and formal discussion on the high-order spatial interactions can be an important future direction.


\paragrapha{Relation to dot-product self-attention. } 
Although the computation of our \gnconv{} largely differs from dot-product self-attention, we will show that \gnconv{} also accomplishes the goal of input-adaptive spatial mixing. Let $\mathbf{M}$ be the attention matrix obtained by multi-head self-attention (MHSA), we write $\mathbf{M}$ as $(m_{ij}^c)$ since the mixing weight may vary across the channels. The spatial mixing result (before the final channel mixing projection) of the $c$-th channel at location $i$ is 
\begin{equation}
  x_{\rm MHSA}^{(i, c)} = \sum_{j\in \Omega_{i}}m_{ij}^cv^{(i,j)}=\sum_{j\in\Omega_i}\sum_{c'=1}^C\underline{m_{ij}^c}w_V^{(c', c)}x^{(j, c')},\label{equ:relation_SA}
\end{equation}
where $w_V$ is the weight of the V-projection layer. Note that $m_{ij}$ obtained by the dot-product operation contains 1-order interaction. On the other hand, the output of our \gnconv{} (before the $\phi_{\rm out}$) can be written as
\begin{equation}
  x_{\gnconv{}}^{(i, c)}=p_n^{(i, c)}=\sum_{j\in\Omega_i}\sum_{c'=1}^C\underline{ w_{n-1, i\to j}^{c}\mathbf{g}_{n-1}^{(i, c)}}w_{\phi_{\rm in}}^{(c', c)}x^{(j, c')}\triangleq \sum_{j\in\Omega_i}\sum_{c'=1}^C\underline{h_{ij}^c}w_{\phi_{\rm in}}^{(c', c)}x^{(j, c')},\label{equ:relation_gnconv}
\end{equation}
where $w_{n-1}$ is the convolutional weight for $f_{n-1}$, $w_{\phi_{\rm in}}$ is the linear weight of $\phi_{\rm in}$, and $\mathbf{g}_{n-1}=g_{n-1}(\mathbf{p}_{n-1})$ is a projection of $\mathbf{p}_{n-1}$. From the formulation in Equation~\eqref{equ:relation_gnconv} we find our \gnconv{} also achieves input-adaptive spatial mixing with $\{h_{ij}^c\}$ as the weights. Observing that $h_{ij}$ is computed from $\mathbf{p}_{n-1}$ which contains $n-1$ order interactions, we can regard our \gnconv{} as an extension of the self-attention in terms of the order of the spatial mixing weight. Therefore, our \gnconv{} can better model more complex spatial interactions. 

The details of \gnconv{} and our implementation are summarized in Figure~\ref{fig:block}.

\vspace{-3pt}
\subsection{Model Architectures}\label{sec:hornet}
\vspace{-3pt}

\paragrapha{HorNet. } The \gnconv{} can be a drop-in replacement of the spatial mixing layer in vision Transformers~\cite{touvron2020deit,liu2021swin} or modern CNNs~\cite{liu2022convnet}. We follow the same meta-architecture as~\cite{Vaswani2017transformer,liu2021swin} to construct HorNet, where the basic block contains a spatial mixing layer and a feed-forward network (FFN). Depending on the model size and the implementation of the depth-wise convolution $f_k$ in our \gnconv{}, we have two series of model variants named HorNet-T/S/B/L$_{7\times 7}$ and HorNet-T/S/B/L$_{\rm GF}$. We consider the popular Swin Transformer~\cite{liu2021swin} and ConvNeXt~\cite{liu2022convnet} as the vision Transformer and CNN baselines since our models are implemented based on a convolution-based framework while having high-order interactions like vision Transformers. To fairly compare with the baselines, we directly follow the number of blocks of Swin Transformers-S/B/L~\cite{liu2021swin} but insert an extra block to the stage 2 to make the overall complexity close, resulting in $[2, 3, 18, 2]$ blocks in each stage in all of the model variants. We simply adjust the base number of channels $C$ to construct models with different sizes and set the number of channels in 4 stages as $[C, 2C, 4C, 8C]$ following common practice. We use $C=64,96,128,192$ for HorNet-T/S/B/L, respectively. We set the interaction orders (\ie, the $n$ in \gnconv{}) for each stage as 2,3,4,5 by default, such that the channels of the coarsest order $C_0$ is the same across different stages. 

\paragrapha{HorFPN. } Apart from using \gnconv{} in visual encoders, we find our \gnconv{} can be an enhanced alternative for standard convolution that considers higher-order spatial interactions in a wide range of convolution-based models. Thus, we replace spatial convolutions for feature fusion in the FPN~\cite{lin2017feature} with our \gnconv{} to improve spatial interactions for downstream tasks. Specifically, we add our \gnconv{} after the fusion of features from different pyramid levels. For object detection, we replace the 3$\times$3 convolution after the top-down pathway with the \gnconv{} in each level. For semantic segmentation, we simply replace the 3$\times$3 convolution after the concatenation of the multi-level feature maps with \gnconv{} since the final results are directly predicted from this concatenated feature. We also have two implementations called HorFPN$_{7\times 7}$ and HorFPN$_{\rm GF}$ decided by the choice of $f_k$.

\vspace{-3pt}
\section{Experiments}
\vspace{-3pt}

We conduct extensive experiments to verify the effectiveness of our method. We present the main results on ImageNet~\cite{deng2009imagenet} and compare them with various architectures. We also test our models on the downstream dense prediction tasks on commonly used semantic segmentation benchmark ADE20K~\cite{zhou2017scene} and object detection dataset COCO~\cite{lin2014coco}. Lastly, we provide ablation studies of our designs and analyze the effectiveness of \gnconv{} on a wide range of models.

\newcommand{\midsepnew}{\aboverulesep = 0.605mm \belowrulesep = 0.984mm}
\newcommand{\midsepdefault}{\aboverulesep = 0.605mm \belowrulesep = 0.984mm}
\begin{table}[t]
 \centering
 \caption{\small \textbf{ImageNet classification results.} We compare our models with state-of-the-art vision Transformers and CNNs that have comparable FLOPs and parameters. We report the top-1 accuracy on the validation set of ImageNet as well as the number of parameters and FLOPs. We also show the improvements over Swin Trasnformers that have similar overall architectures and training configurations to our models. “↑384” indicates that the model is fine-tuned on 384$\times$384 images for 30 epochs. Our models are highlighted in \hl{gray}. }
 
 \begin{minipage}{.49\textwidth}
 \adjustbox{width=\linewidth}{
 \midsepnew
 \setlength{\tabcolsep}{2pt}
  \begin{tabular}{L{120pt}L{35pt}L{30pt}L{30pt}L{40pt}}\toprule
  \multirow{2}[2]{*}{Model} & Image & Params & FLOPs & Top-1 \\ 
  & Size & (M) & (G) & Acc. (\%) \\
  \midrule
  \multicolumn{5}{l}{\emph{ImageNet-1K trained models}} \\
  \midrule
  EfficientNet-B4~\cite{tan2019efficientnet} & $380^2$ & 19 & 4.2 & 82.9 \\
  EfficientNet-B5~\cite{tan2019efficientnet} & $456^2$ & 30 & 9.9 & 83.6 \\ 
  EfficientNet-B6~\cite{tan2019efficientnet} & $528^2$ & 43 & 19.0 & 84.0 \\ 
  EfficientNetV2-S~\cite{tan2019efficientnet} & $300^2$ & 24 & 8.8 & 83.9\\
  RepLKNet-31B~\cite{ding2022scaling} & $224^2$ &79 & 15.3 & 83.5 \\
  VAN-B~\cite{guo2022van} & $224^2$ &27 & 5.0 & 82.8\\
  VAN-L~\cite{guo2022van} & $224^2$ &45 & 9.0 & 83.9\\
  CSWin-T~\cite{dong2021cswin} & $224^2$ &23 & 4.3 & 82.7 \\
  CSWin-S~\cite{dong2021cswin} & $224^2$ &35 & 6.9 & 83.6 \\
  CSWin-B~\cite{dong2021cswin} & $224^2$ &78 & 15.0 & 84.2 \\
  \midrule
  Swin-T~\cite{liu2021swin} & $224^2$ &28 & 4.5 & 81.3 \\
  ConvNeXt-T~\cite{liu2022convnet} & $224^2$ &29 & 4.5 & 82.1$_{\cb{\text{(+0.7)}}}$ \\
  \rowcolor{Gray} HorNet-T$_{7\times 7}$ & $224^2$ &22 & 4.0 & 82.8$_{\cb{\text{(+1.5)}}}$ \\
  \rowcolor{Gray} HorNet-T$_{\rm GF}$ & $224^2$ &23 & 3.9 & \textbf{83.0}$_{\cb{\text{(+1.7)}}}$ \\
   \midrule
  Swin-S~\cite{liu2021swin} & $224^2$ &50 & 8.7 & 83.0 \\
  ConvNeXt-S~\cite{liu2022convnet} & $224^2$ &50 & 8.7 & 83.1$_{\cb{\text{(+0.1)}}}$ \\
  \rowcolor{Gray} HorNet-S$_{7\times 7}$ & $224^2$ &50 & 8.8 & 83.8$_{\cb{\text{(+0.8)}}}$ \\
  \rowcolor{Gray} HorNet-S$_{\rm GF}$ & $224^2$ &50 & 8.7 & \textbf{84.0}$_{\cb{\text{(+1.0)}}}$ \\
   \midrule
  Swin-B~\cite{liu2021swin} & $224^2$ &89 & 15.4 & 83.5 \\
  ConvNeXt-B~\cite{liu2022convnet} & $224^2$ &88 & 15.4 & 83.8$_{\cb{\text{(+0.3)}}}$ \\
  \rowcolor{Gray}HorNet-B$_{7\times 7}$ & $224^2$ &87 & 15.6 & 84.2$_{\cb{\text{(+0.7)}}}$ \\
  \rowcolor{Gray}HorNet-B$_{\rm GF}$ & $224^2$ &88 & 15.5 & \textbf{84.3}$_{\cb{\text{(+0.8)}}}$ \\
  \bottomrule
   \end{tabular}%
   \midsepdefault
  }
  \end{minipage}
  \hfill
  \begin{minipage}{.49\textwidth}
 \adjustbox{width=\linewidth}{
 \midsepnew
 \setlength{\tabcolsep}{2pt}
  \begin{tabular}{L{120pt}L{35pt}L{30pt}L{30pt}L{40pt}}\toprule
  \multirow{2}[2]{*}{Model} & Image & Params & FLOPs & Top-1 \\ 
  & Size & (M) & (G) & Acc. (\%) \\
  \midrule
  \multicolumn{5}{l}{\emph{ImageNet-1K trained models (fine-tuned at 384$\mathit{\times}$384) }} \\
  \midrule
   Swin-B↑384~\cite{liu2021swin} & $384^2$ & 89 & 47.1 & 84.5 \\
  ConvNeXt-B↑384~\cite{liu2022convnet} & $384^2$ &88 & 45.0 & 85.1$_{\cb{\text{(+0.6)}}}$ \\
  \rowcolor{Gray}HorNet-B$_{7\times 7}$↑384 & $384^2$ &87 & 45.8 & 85.3$_{\cb{\text{(+0.8)}}}$ \\
  \rowcolor{Gray}HorNet-B$_{\rm GF}$↑384 & $384^2$ &92 & 45.4 & \textbf{85.6}$_{\cb{\text{(+1.1)}}}$ \\
  \bottomrule\addlinespace[9.2pt]
  \multicolumn{5}{l}{\emph{ImageNet-22K trained models (fine-tuned to ImageNet-1K) }} \\
  \midrule
  R-101x3~\cite{kolesnikov2020big} & $384^2$ & 388 & 204.6 & 84.4 \\
  R-152x4~\cite{kolesnikov2020big} & $480^2$ & 937 & 840.5 & 85.4 \\
  ViT-B/16~\cite{dosovitskiy2020vit} & $384^2$ &87 & 55.5 & 84.0 \\
  ViT-L/16~\cite{dosovitskiy2020vit} & $384^2$ &305 & 191.1 & 85.2 \\ 
  
  EfficientNetV2-L~\cite{tan2019efficientnet} & $380^2$ & 121 & 53.0 & 86.8 \\ 
  CSWin-L~\cite{dong2021cswin} & $384^2$ & 173 & 96.8 & 87.5 \\
  SwinV2-L~\cite{liu2021swinv2} & $384^2$ & 197 & 115.4 & 87.6 \\
  RepLKNet-31L~\cite{ding2022scaling} & $384^2$ & 172 & 96.0 & 86.6 \\ 
  \midrule
  Swin-L~\cite{liu2021swin} & $224^2$ & 197 & 34.5 & 86.3 \\
  ConvNeXt-L~\cite{liu2022convnet} & $224^2$ & 198 & 34.4 & 86.6$_{\cb{\text{(+0.3)}}}$ \\
  \rowcolor{Gray}HorNet-L$_{7\times 7}$ & $224^2$ & 195 & 34.8 & 86.8$_{\cb{\text{(+0.5)}}}$ \\
  \rowcolor{Gray}HorNet-L$_{\rm GF}$ & $224^2$ & 196 & 34.6 & \textbf{87.0}$_{\cb{\text{(+0.7)}}}$ \\
  \midrule
  Swin-L↑384~\cite{liu2021swin} & $384^2$ & 197 & 103.9 & 87.3 \\
  ConvNeXt-L↑384~\cite{liu2022convnet} & $384^2$ & 198 & 101.0 & 87.5$_{\cb{\text{(+0.2)}}}$ \\
  \rowcolor{Gray}HorNet-L$_{7\times 7}$↑384 & $384^2$ & 195 & 102.3 & 87.6$_{\cb{\text{(+0.3)}}}$ \\
  \rowcolor{Gray}HorNet-L$_{\rm GF}$↑384 & $384^2$ & 202 & 101.8 &  \textbf{87.7}$_{\cb{\text{(+0.4)}}}$ \\
  \bottomrule
   \end{tabular}%
   \midsepdefault
  }
  \end{minipage}
  
 \label{tab:imagenet}%
  \vspace{-10pt}
\end{table}%

\vspace{-3pt}
\subsection{ImageNet Classification}
\vspace{-3pt}

\paragrapha{Setups.} We conduct image classification experiments on the widely used ImageNet~\cite{deng2009imagenet} dataset. We train our HorNet-T/S/B models using the standard ImageNet-1K dataset following common practice. To fairly compare with previous work, we directly use the training configurations of~\cite{liu2022convnet,liu2021swin,touvron2020deit} to train our models. We train the models for 300 epochs with $224\times 224$ input. 
To evaluate the scaling ability of our designs, we further train the HorNet-L models on the ImageNet-22K dataset that contains over $10\times$ images and more categories. We follow previous practice~\cite{liu2021swin, liu2022convnet} to train our models for 90 epochs and use a similar data augmentation strategy as ImageNet-1K experiments. We fine-tune the models pre-trained on ImageNet-22K or at the 224×224
resolution to ImageNet-1K or/and 384×384 resolution for 30 epochs following~\cite{liu2022convnet}. When adapting the ImageNet-22K models to ImageNet-1K, we initialize the classifier with the pre-trained class centers to stabilize the training process.  More details can be found in Appendix~\ref{appendix:train_details}.

\paragrapha{Results.} The results of our ImageNet classification experiments are summarized in Table~\ref{tab:imagenet}. We see that our models achieve very competitive performance with state-of-the-art vision Transformers and CNNs. Notably, HorNet surpasses Swin Transformers and ConvNeXt which have similar overall architectures and training configurations by a healthy margin on various model sizes and settings.  Our models also generalize well to a larger image resolution, larger model sizes and more training data. These results clearly demonstrate the effectiveness and generality of our designs.

\vspace{-3pt}
\subsection{Dense Prediction Tasks}
\vspace{-3pt}

\begin{table}[t]
 \centering 
 \caption{\small \textbf{Object detection and semantic segmentation results with different backbones.} We use UperNet~\cite{xiao2018unified} for semantic segmentation and Cascade Mask R-CNN~\cite{cai2018cascade} for object detection. $^\ddagger$ indicates that the model is pre-trained on ImageNet-22K. For semantic segmentation, we report both single-scale (SS) and multi-scale (MS) mIoU. The FLOPs are calculated with image size (2048, 512) for ImageNet-1K pre-trained models and (2560, 640) for ImageNet-22K pre-trained models. For object detection, we report the box AP and the mask AP. FLOPs are measured on input sizes of (1280, 800). Our models are highlighted in \hl{gray}.}
 \adjustbox{width=\linewidth}{
  \begin{tabular}{lC{45pt}C{45pt}C{45pt}C{45pt}C{45pt}C{45pt}C{45pt}C{45pt}}
  \toprule 
  \multicolumn{1}{l}{\multirow{2}[3]{*}{Backbone}} & \multicolumn{4}{c}{Semantic Segmentation with \emph{UperNet 160K}}  & \multicolumn{4}{c}{Object Detection with \emph{Cascade Mask R-CNN 3$\times$}} \\
  \cmidrule(lr){2-5}\cmidrule(lr){6-9}
     & mIoU$^{\rm ss}$ & mIoU$^{\rm ms}$ & Params & FLOPs & AP$^{\rm box}$ & AP$^{\rm mask}$ & Params & FLOPs \\
  \midrule
  Swin-T~\cite{liu2021swin}~~~~~~~~~~~~~~~~~~~~~~~~~~ & 44.5 & 45.8 & 60M  & 945G & 50.4 & 43.7 & 86M  & 745G \\
  ConvNeXt-T~\cite{liu2022convnet} & 46.0 & 46.7 & 60M  & 939G & 50.4 & 43.7 & 86M  & 741G \\
  \rowcolor{Gray} HorNet-T$_{7\times 7}$ & 48.1  & 48.9  & 52M   &  926G  &   51.7 & 44.8  & 80M & 730G \\
  \rowcolor{Gray} HorNet-T$_{\rm GF}$ & \textbf{49.2} & \textbf{49.3} &  55M  &  924G  &  \textbf{52.4}  & \textbf{45.6}  & 80M & 728G\\\midrule
  Swin-S~\cite{liu2021swin} & 47.6 & 49.5 & 81M  & 1038G & 51.9 & 45.0 & 107M & 838G \\
  ConvNeXt-S~\cite{liu2022convnet} & 48.7 & 49.6 & 82M  & 1027G & 51.9 & 45.0 & 108M & 827G \\
  \rowcolor{Gray} HorNet-S$_{7\times 7}$ & 49.2 & 49.8 & 81M  & 1030G & 52.7 & 45.6 & 107M & 830G \\
  \rowcolor{Gray} HorNet-S$_{\rm GF}$ & \textbf{50.0} & \textbf{50.5} & 85M  & 1027G & \textbf{53.3} & \textbf{46.3} & 108M & 827G \\\midrule
  Swin-B~\cite{liu2021swin} & 48.1 & 49.7 & 121M & 1188G & 51.9 & 45.0 & 145M & 982G \\
  ConvNeXt-B~\cite{liu2022convnet} & 49.1 & 49.9 & 122M & 1170G & 52.7 & 45.6 & 146M & 964G \\
  \rowcolor{Gray} HorNet-B$_{7\times 7}$ & 50.0 & 50.5  & 121M & 1174G & 53.3 & 46.1 & 144M & 969G \\
  \rowcolor{Gray} HorNet-B$_{\rm GF}$ & \textbf{50.5} & \textbf{50.9} & 126M & 1171G & \textbf{54.0} & \textbf{46.9} & 146M & 965G \\\midrule\midrule
  Swin-L$^\ddagger$~\cite{liu2021swin} & 52.1 & 53.5 & 234M & 2468G & 53.9 & 46.7 & 253M & 1382G \\
  ConvNeXt-L$^\ddagger$~\cite{liu2022convnet} & 53.2 & 53.7 & 235M & 2458G & 54.8 & 47.6 & 255M & 1354G \\
 \textcolor{gray}{ConvNeXt-XL$^\ddagger$~\cite{liu2022convnet}} &  \textcolor{gray}{53.6} &  \textcolor{gray}{54.0} &  \textcolor{gray}{391M} &  \textcolor{gray}{3335G} &  \textcolor{gray}{55.2} &  \textcolor{gray}{47.7} &  \textcolor{gray}{407M} & \textcolor{gray}{1898G} \\ 
  \rowcolor{Gray} HorNet-L$^\ddagger_{7\times 7}$ & 54.1 & 54.5 & 232M & 2473G & 55.4 & 48.0 & 251M & 1363G \\
  \rowcolor{Gray} HorNet-L$^\ddagger_{\rm GF}$ & \textbf{55.0} & \textbf{55.2} & 239M & 2465G & \textbf{56.0} & \textbf{48.6} & 259M & 1358G \\\bottomrule
  \end{tabular}%
  }
   \vspace{-10pt}

 \label{tab:downstream}%
\end{table}%

\begin{table}[t]
 \centering
 \caption{\small\textbf{Comparisons of HorFPN with standard FPN on different backbones. } We use UperNet 160K and Mask R-CNN 1$\times$ schedule for semantic segmentation and object detection, respectively. We find our HorFPN consistently outperforms standard FPN with various of backbones on both the two tasks.}
 \adjustbox{width=\linewidth}{
  \begin{tabular}{llC{45pt}C{45pt}C{45pt}C{45pt}C{45pt}C{45pt}C{45pt}C{45pt}}\toprule
  \multirow{2}[0]{*}{Backbone} & \multicolumn{1}{l}{\multirow{2}[1]{*}{\makecell[c]{Fusion\\ Module}}} & \multicolumn{4}{c}{Semantic Segmentation with \textit{UperNet 160K}} & \multicolumn{4}{c}{Object Detection with \textit{Mask R-CNN 1}$\times$} \\\cmidrule(lr){3-6}\cmidrule(lr){7-10}
     &    & \multicolumn{1}{c}{mIoU$^{\rm ss}$} & mIoU$^{\rm ms}$ & \multicolumn{1}{c}{Params} & \multicolumn{1}{c}{FLOPs} & AP$^{\rm box}$ & AP$^{\rm mask}$ & \multicolumn{1}{c}{Params} & \multicolumn{1}{c}{FLOPs} \\\midrule
  \multirow{3}[0]{*}{ResNet-50~\cite{he2016deep}} & FPN~\cite{lin2017feature}  & 40.7 & 41.8 & 66M  & 947G & 38.2 & 34.7 & 44M  & 260G \\
     & HorFPN$_{7\times 7}$ & 41.8 & 44.1 & 60M  & 499G & 38.7 & 35.1 & 43M  & 226G \\
     & HorFPN$_{\rm GF}$ & \textbf{43.2} & \textbf{44.5} & 60M  & 497G & \textbf{39.1} & \textbf{35.5} & 43M  & 224G \\\midrule
  \multirow{3}[0]{*}{ResNet-101~\cite{he2016deep}} & FPN~\cite{lin2017feature}  & 42.9 & 44.0 & 85M  & 1025G & 40.0 & 36.1 & 63M  & 336G \\
     & HorFPN$_{7\times 7}$ & 44.1 & 45.5 & 79M  & 577G & 40.3 & 36.4 & 62M  & 302G \\
     & HorFPN$_{\rm GF}$ & \textbf{44.5} & \textbf{46.4} & 79M  & 574G & \textbf{40.5} & \textbf{36.7} & 62M  & 300G \\\midrule
  \multirow{3}[0]{*}{Swin-S~\cite{liu2021swin}} & FPN~\cite{lin2017feature}  & 47.6 & 49.5 & 81M  & 1038G & 45.5 & 40.9 & 69M  & 354G \\
     & HorFPN$_{7\times 7}$ & 48.0 & 49.2 & 74M  & 580G & 46.3 & 41.1 & 68M  & 325G \\
     & HorFPN$_{\rm GF}$ & \textbf{49.0} & \textbf{49.9} & 75M  & 578G & \textbf{46.8} & \textbf{41.9} & 69M  & 323G \\\midrule
  \multirow{3}[0]{*}{HorNet-S} & FPN~\cite{lin2017feature}  & 49.2 & 49.8 & 81M  & 1030G & 47.1 & 42.2 & 69M  & 351G \\
     & HorFPN$_{7\times 7}$ & 49.4 & 50.1 & 74M  & 577G & 47.4 & 42.3 & 68M  & 322G \\
     & HorFPN$_{\rm GF}$ & \textbf{49.7} & \textbf{50.3} & 75M  & 575G & \textbf{47.7} & \textbf{42.4}   & 68M  & 321G \\\bottomrule
  \end{tabular}%
  } \vspace{-10pt}
 \label{tab:horfpn}%
\end{table}%

\paragrapha{HorNet for semantic segmentation. } We evaluate our HorNet for semantic segmentation task on ADE20K~\cite{zhou2017scene} dataset using the commonly used UperNet~\cite{xiao2018unified} framework. All the models are trained for 160k iterations using AdamW~\cite{adamw} optimizer with a global batch size of 16. The image size during training is $512\times 512$ for ImagNet-1k (HorNet-T/S/B) pre-trained models and $640\times 640$ for the ImageNet-22K pre-trained models (HorNet-L). The results are summarized in the left part of Table~\ref{tab:downstream}, where we report both the single-scale (SS) and multi-scale (MS) mIoU on the validation set. Both our HorNet$_{7\times 7}$ and HorNet$_{\rm GF}$ models outperform Swin~\cite{liu2021swin} and ConvNeXt~\cite{liu2022convnet} models with similar model sizes and FLOPs. Specifically, HorNet$_{\rm GF}$ models achieve better results than HorNet$_{\rm 7\times 7}$ and ConvNeXt series by large margins in single-scale mIoU, indicating the global interactions captured by the global filter are helpful for semantic segmentation. Notably, we find both our HorNet-L$_{7\times 7}$ and HorNet-L$_{\rm GF}$ even outperform ConvNeXt-XL with $\sim$25\% fewer FLOPs. These results clearly demonstrate the effectiveness and scalability of our HorNet on semantic segmentation.

\begin{wraptable}{R}{0.6\linewidth}
 \centering 
 \vspace{-12pt}
 \caption{\small \textbf{Object detection results with recent state-of-the-art frameworks.} We report the single-scale AP$^{\rm box}$ and AP$^{\rm mask}$  on the validation set of COCO. Our models are highlighted in \hl{gray}.}  \label{tab:sotadet}%
 \adjustbox{width=\linewidth}{
  \begin{tabular}{C{90pt}C{80pt}C{50pt}C{50pt}}
  \toprule 
  Backbone &  Framework & AP$^{\rm box}$ & AP$^{\rm mask}$ \\ \midrule
  Swin-L~\cite{liu2021swin} & HTC++~\cite{chen2019hybrid} & 57.1 & 49.5 \\
  ViT-Adapter-L~\cite{chen2022vision} & HTC++~\cite{chen2019hybrid} & 57.9 & 50.2 \\
  \rowcolor{Gray} HorNet-L$_{\rm GF}$ & HTC++~\cite{chen2019hybrid} & \textbf{58.1} & \textbf{50.5} \\ \midrule
    Swin-L~\cite{liu2021swin} & DINO~\cite{zhang2022dino} & 58.5 & - \\
    \rowcolor{Gray} HorNet-L$_{\rm GF}$ & DINO~\cite{zhang2022dino} & \textbf{59.2} & - \\
  \bottomrule
  \end{tabular}%
  }
  
  \caption{\small \textbf{Semantic Segmentation results with recent state-of-the-art frameworks.} We report the single-scale (SS) and multi-scale (MS) mIoU on the validation set of ADE20K. Our models are highlighted in \hl{gray}.} \label{tab:sotaseg}%
 \adjustbox{width=\linewidth}{
  \begin{tabular}{C{90pt}C{80pt}C{50pt}C{50pt}}
  \toprule 
  Backbone &  Framework & mIoU$^{\rm ss}$ & mIoU$^{\rm ms}$ \\ \midrule
  Swin-L~\cite{liu2021swin} & Mask2Former~\cite{cheng2022masked} & 56.1 & 57.3 \\
  Swin-L-FaPN~\cite{huang2021fapn}  & Mask2Former~\cite{cheng2022masked} & 56.4 & 57.7 \\
  \rowcolor{Gray} HorNet-L$_{\rm GF}$ & Mask2Former~\cite{cheng2022masked} & \textbf{57.5} & \textbf{57.9} \\ 
  \bottomrule
  \end{tabular}%
  }
  \vspace{-12pt}
\end{wraptable}%

\paragrapha{HorNet for object detection. } We also evaluate our models on the COCO~\cite{lin2014coco} dataset. We adopt the cascade Mask R-CNN framework~\cite{he2017mask,cai2018cascade} to perform object detection and instance segmentation using HorNet-T/S/B/L backbones. Following Swin~\cite{liu2021swin} and ConvNeXt~\cite{liu2022convnet}, we use $3\times$ schedule with multi-scale training. The right part of Table~\ref{tab:downstream} compares the box AP and mask AP of our HorNet models and Swin/ConvNeXt models. Similarly, we show our HorNet models achieve consistently and significantly better performance than the Swin/ConvNeXt counterparts, in both box AP and mask AP. The HorNet$_{\rm GF}$ series obtain +1.2$\sim$2.0 box AP and +1.0$\sim$1.9 mask AP compared with ConvNeXt. Again, our large model HorNet-L$_{7\times 7}$ and HorNet$_{\rm GF}$ can outperform ConvNeXt-XL, which further validates the favorable transferability with a larger model size and larger pre-trained dataset.

\paragrapha{HorFPN for dense prediction. } We now show another application of the proposed \gnconv{}, \ie, to serve as a better fusion module that can better capture the higher-order interactions among different levels of features in dense prediction tasks. Specifically, we directly modify the FPN~\cite{lin2017feature} as described in Section~\ref{sec:hornet} in UperNet~\cite{xiao2018unified} and Mask R-CNN~\cite{he2017mask} for semantic segmentation and object detection, respectively.
We show the results in Table~\ref{tab:horfpn}, where we compare the performance of our HorFPN and standard FPN on different backbones including ResNet-50/101~\cite{he2016deep}, Swin-S~\cite{liu2021swin} and HorNet-S$_{7\times 7}$. For semantic segmentation, we find our HorFPN can significantly reduce the FLOPs ($\sim$50\%) while achieving better validation mIoU. For object detection, our HorFPN can also outperform standard FPN in terms of both box AP and mask AP on different backbones with about 30G fewer FLOPs. Besides, we observe that the HorFPN$_{\rm GF}$ is consistently better than HorFPN$_{\rm 7\times 7}$, indicating that global interactions are also important when fusing hierarchical features.

\paragrapha{Results with state-of-the-art frameworks. } \label{appendix:sota} To further show the effectiveness our backbone, we conduct experiments to combine our large HorNet model with recent state-of-the-art dense prediction frameworks including HTC++~\cite{chen2019hybrid}, DINO~\cite{zhang2022dino} and Mask2Former~\cite{cheng2022masked}. For HTC++ and DINO, we train our models on COCO for 36 epochs (3$\times$ schedule) and does not introduce extra pre-training data like Object365 in~\cite{zhang2022dino}. We report the single-scale performance on the validation set and compared with several state-of-the-art methods in Table~\ref{tab:sotadet}. For Mask2Former, we train our models on ADE20K with $640\times 640$. We report the mIoU of both single-scale and multi-scale testing on the validation set in Table~\ref{tab:sotaseg}.

\begin{table}
\caption{\textbf{Ablation study and results of applying \gnconv{} to other models/operations. } We provide the ablation study of our designs in (a). [*] indicates the baseline of our model. The baseline and our final models are highlighted in \hl{gray}. In (b) and (c), we apply the proposed \gnconv{} to isotropic models that have a similar level of complexity with ViT/DeiT-S~\cite{dosovitskiy2020vit,touvron2020deit} and other spatial mixing operations including the 3$\times$3 depth-wise convolution and 3$\times$3 pooling used in~\cite{yu2021metaformer}. }\label{tab:analysis} \vspace{-5pt}
\begin{minipage}{.57\textwidth}
\begin{subtable}{\linewidth}
\caption{\footnotesize Ablation study. }\label{tab:abl:model}
\vspace{-5pt}
 \adjustbox{width=\linewidth}{
  \begin{tabular}{llll}\toprule
  Model & Params & FLOPs & Acc. (\%) \\ \midrule
  Swin-T~\cite{liu2021swin} & 28M & 4.5G & 81.3$_{\cb{\text{(+0.1)}}}$ \\ 
  \midrule
  \rowcolor{Gray} - Self-Attention + DWConv$_{7\times 7}$ [*] & 29M & 4.5G & 81.2 \\
  + SE~\cite{senet} & 30M & 4.5G & 81.5$_{\cb{\text{(+0.3)}}}$ \\ 
  - SE + $\textit{g}^\text{\{1,1,1,1\}}$Conv & 28M & 4.3G & 81.7$_{\cb{\text{(+0.5)}}}$\\
  + $\textit{g}^\text{\{2,2,2,2\}}$Conv &28M & 4.3G & 82.2$_{\cb{\text{(+1.0)}}}$ \\
  + $\textit{g}^\text{\{3,3,3,3\}}$Conv & 28M& 4.3G & 82.5$_{\cb{\text{(+1.3)}}}$ \\
  + $\textit{g}^\text{\{4,4,4,4\}}$Conv & 28M& 4.3G & 82.5$_{\cb{\text{(+1.3)}}}$ \\
  + $\textit{g}^\text{\{1,2,3,4\}}$Conv & 28M& 4.3G & 82.5$_{\cb{\text{(+1.3)}}}$ \\
  + $\textit{g}^\text{\{2,3,4,5\}}$Conv & 28M& 4.3G & 82.6$_{\cb{\text{(+1.4)}}}$ \\
  \rowcolor{Gray}+ Deeper \& Narrower (HorNet-T$_{7\times 7}$) & 22M & 4.0G & 82.8$_{\cb{\text{(+1.6)}}}$ \\
  \rowcolor{Gray}+ Global Filters~\cite{rao2021global} (HorNet-T$_{\rm GF}$) & 23M & 3.9G & 83.0$_{\cb{\text{(+1.8)}}}$ \\
  \midrule
  ConvNeXt~\cite{liu2022convnet} & 28M & 4.5G & 82.1$_{\cb{\text{(+0.9)}}}$ \\
  \bottomrule
  \end{tabular}%
}

\end{subtable}
\end{minipage}%
\hfill
\begin{minipage}{.4\textwidth}
\begin{subtable}{\linewidth}
\caption{\footnotesize Results on isotropic models.}\label{tab:abl:iso}
\vspace{-5pt}
\adjustbox{width=\linewidth}{
  \begin{tabular}{lcc} \toprule
   Model & FLOPs& Acc. (\%) \\
  \midrule
  DeiT-S~\cite{touvron2020deit} & 4.6G & 79.8 \\
  ConvNeXt-S (iso.)~\cite{liu2022convnet} & 4.3G & 79.7 \\
  \rowcolor{Gray} HorNet-S$_{7\times 7}$ (iso.) & 4.5G & 80.6\\
  \rowcolor{Gray} HorNet-S$_{\rm GF}$ (iso.) & 4.5G & 81.0 \\
  \bottomrule
  \end{tabular}
}

\end{subtable}
\vspace{1pt}
\begin{subtable}{\linewidth}
\caption{\footnotesize \gnconv{} for other operations. }\label{tab:abl:other}
\vspace{-5pt}
\adjustbox{width=\linewidth}{
  \begin{tabular}{lcc} \toprule
   Model~~~~~~~~~~~~~~~~~~~~~~~~~~~~ & FLOPs& Acc. (\%) \\
  \midrule
  DWConv$_{3\times 3}$ & 4.0G & 80.7 \\
   \rowcolor{Gray} \gnconv{}$_{3\times 3}$ & 3.9G & 82.1 \\
  Pool~\cite{yu2021metaformer} & 3.9G & 78.1 \\
  \rowcolor{Gray} \gnconv{}$_{\rm pool}$ & 3.8G & 79.3 \\
  \bottomrule
  \end{tabular}
}
\end{subtable}
\end{minipage}

\end{table}

\begin{figure}
  \centering
  \includegraphics[width=\linewidth]{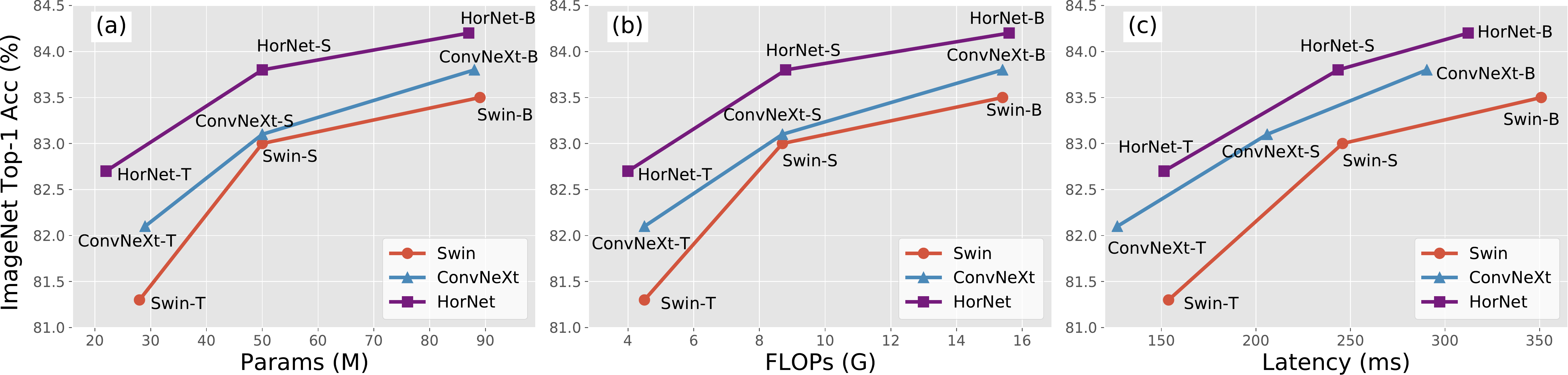}
  \caption{\small \textbf{Comparisons of trade-offs of Swin, ConvNeXt and HorNet.} We compare the trade-offs of the models via the top-1 accuracy on ImageNet \wrt \textbf{(a)} number of parameters; \textbf{(b)} FLOPs; \textbf{(c)} latency. The latency is measured with a single NVIDIA RTX 3090 GPU with a batch size of 128.} \vspace{-10pt}
  \label{fig:tradeoff}
\end{figure}

\vspace{-3pt}
\subsection{Analysis} \label{sec:analysis}
\vspace{-3pt}

\paragrapha{Ablation study.} We provide detailed ablation studies of the \gnconv{} and our HorNet in Table~\ref{tab:analysis}. We first study the model designs of our HorNet in Table~\ref{tab:abl:model}. Our baseline ([*]) is obtained by simply replacing the self-attention with 7$\times$7 depth-wise convolution in Swin-T~\cite{liu2021swin}. We first show that both SE~\cite{senet} and our \gnconv{} with $n=1$ ($\textit{g}^\text{\{1,1,1,1\}}$Conv) can improve over the baseline model [*], and $\textit{g}^\text{\{1,1,1,1\}}$Conv is slightly better. We then perform ablations on the interaction order $n$ for each stage and find: (1) if $n$ is shared across the 4 stages, the accuracy will increase with larger $n$ but saturate at 82.5 when $n=4$; (2) progressively increased order ($\textit{g}^\text{\{2,3,4,5\}}$Conv) can further improve the accuracy. Our final models are built on $\textit{g}^\text{\{2,3,4,5\}}$Conv by adjusting the depth and width of the networks (HorNet-T$_{7\times 7}$) and applying Global Filter~\cite{rao2021global} for the depth-wise convolution (HorNet-T$_{\rm GF}$). These results clearly show that our \gnconv{} is an efficient and extendable operation that can better capture high-order spatial interactions than both self-attention and depth-wise convolution.

\paragrapha{\gnconv{} for isotropic models.} We also evaluate \gnconv{} on isotropic architectures (with constant spatial resolutions). We replace the self-attention in DeiT-S~\cite{touvron2020deit} with our \gnconv{} and adjust the number of blocks to 13 to obtain the isotropic HorNet-S$_{\rm 7\times 7}$ and HorNet-S$_{\rm GF}$. We compare DeiT-S, isotropic ConvNeXt-S and isotropic HorNet-S in Table~\ref{tab:abl:iso}. 
While isotropic ConvNeXt-S cannot improve DeiT-S, our isotropic HorNet surpasses DeiT-S by a large margin. These results indicate that our \gnconv{} can better realize the functions of self-attention compared to plain convolutions and  have better ability to model the complex spatial interactions.

\paragrapha{\gnconv{} for other operations.} To further demonstrate the universality of \gnconv{}, we use 3$\times$3 depth-wise convolution and 3$\times$3 pooling~\cite{yu2021metaformer} as the basic operation in the \gnconv{}. The results in Table~\ref{tab:abl:other} show that \gnconv{} can also improve these two operations by large margins, indicating our \gnconv{} is potentially more powerful when equipped with some better basic operations.

\paragrapha{Accuracy-complexity trade-offs.} We visualize accuracy-complexity trade-offs of Swin, ConvNeXt and HorNet series in Figure~\ref{fig:tradeoff}. For fair comparisons, we fix the input image size to $224\times 224$ and use HorNet$_{\rm 7\times 7}$ such that all the compared models are based on 7$\times$7 local window. We see HorNet can achieve better trade-offs than the representative vision Transformers and modern CNNs with regards to model size, FLOPs and GPU latency. 

\begin{figure}[t]
    \centering
    \includegraphics[width=\textwidth]{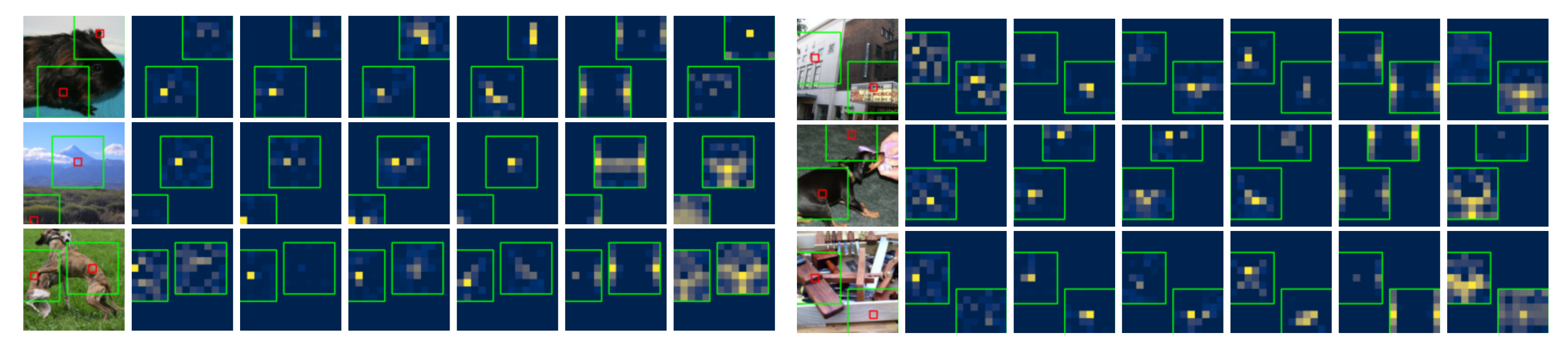}
    \caption{\small Visualization of the adaptive weights generated by \gnconv{}. We see that the spatial mixing weights of our \gnconv{} are adaptive both to input samples and spatial locations, which further indicates that \gnconv{} shares these two desirable characteristics with the self-attention operation.}
    \label{fig:viz}
    \vspace{-10pt}
\end{figure}

\paragrapha{Visualization. } We provide some visualizations of the adaptive weights learned by \gnconv{} in Figure~\ref{fig:viz}. For each sample, we show the value of $\frac{1}{C}\sum_{c=1}^{C}h_{ij}^c$ (see Equation (3.8) or the definition of $h_{ij}^c$) for two random spatial locations $i$ from layer \{1, 3, 5, 7, 8, 12\} of the isotropic HorNet-S model. Figure~\ref{fig:viz} demonstrates that the spatial mixing weights of our \gnconv{} are adaptive both to input samples and spatial locations, which further indicates that \gnconv{} shares these two desirable characteristics with the self-attention operation.

\paragrapha{Limitations.} While HorNet shows better overall latency-accuracy trade-offs, we notice that HorNet is slower than ConvNeXt with similar FLOPs on GPU, which may be caused by the more complex designs to perform the high-order interactions. We think that developing a more hardware-friendly operation for high-order spatial interactions is an interesting future direction to improve our work.

\vspace{-3pt}
\section{Conclusion}
\vspace{-3pt}
\label{sec:con}

 We have presented the Recursive Gated Convolution ($\textit{g}^\textit{n}$Conv) that performs efficient, extendable, and translation-equivariant high-order spatial interactions with gated convolutions and recursive deigns. $\textit{g}^\textit{n}$Conv can serve as a drop-in replace of the spatial mixing layer in various vision Transformers and convolution-based models. Based on the operation, we have constructed a new family of generic vision backbones HorNet. Extensive experiments demonstrate the effectiveness of $\textit{g}^\textit{n}$Conv and HorNet on commonly used visual recognition benchmarks. We hope our attempt can inspire future work to further explore the high-order spatial interactions in vision models. 
 
\section*{Acknowledgments}

Jiwen Lu was supported in part by the National Key Research and Development Program of China under Grant 2017YFA0700802, the National Natural Science Foundation of China under Grant 62125603 and Grant U1813218, and a grant from the Beijing Academy of Artificial Intelligence (BAAI). 

{\small 
\bibliographystyle{plain}
\bibliography{ref}

\begin{thebibliography}{10}

\bibitem{ai2003interaction}
Chunrong Ai and Edward~C Norton.
\newblock Interaction terms in logit and probit models.
\newblock {\em Economics letters}, 80(1):123--129, 2003.

\bibitem{cai2018cascade}
Zhaowei Cai and Nuno Vasconcelos.
\newblock Cascade r-cnn: Delving into high quality object detection.
\newblock In {\em CVPR}, pages 6154--6162, 2018.

\bibitem{chen2019hybrid}
Kai Chen, Jiangmiao Pang, Jiaqi Wang, Yu~Xiong, Xiaoxiao Li, Shuyang Sun,
  Wansen Feng, Ziwei Liu, Jianping Shi, Wanli Ouyang, et~al.
\newblock Hybrid task cascade for instance segmentation.
\newblock In {\em CVPR}, pages 4974--4983, 2019.

\bibitem{chen2020dynamic}
Yinpeng Chen, Xiyang Dai, Mengchen Liu, Dongdong Chen, Lu~Yuan, and Zicheng
  Liu.
\newblock Dynamic convolution: Attention over convolution kernels.
\newblock In {\em CVPR}, pages 11030--11039, 2020.

\bibitem{chen2022vision}
Zhe Chen, Yuchen Duan, Wenhai Wang, Junjun He, Tong Lu, Jifeng Dai, and
  Yu~Qiao.
\newblock Vision transformer adapter for dense predictions.
\newblock {\em arXiv preprint arXiv:2205.08534}, 2022.

\bibitem{cheng2021masked}
Bowen Cheng, Ishan Misra, Alexander~G Schwing, Alexander Kirillov, and Rohit
  Girdhar.
\newblock Masked-attention mask transformer for universal image segmentation.
\newblock {\em arXiv preprint arXiv:2112.01527}, 2021.

\bibitem{cheng2022masked}
Bowen Cheng, Ishan Misra, Alexander~G Schwing, Alexander Kirillov, and Rohit
  Girdhar.
\newblock Masked-attention mask transformer for universal image segmentation.
\newblock In {\em CVPR}, pages 1290--1299, 2022.

\bibitem{cheng2021maskformer}
Bowen Cheng, Alex Schwing, and Alexander Kirillov.
\newblock Per-pixel classification is not all you need for semantic
  segmentation.
\newblock {\em NeurIPS}, 34, 2021.

\bibitem{chu2021twins}
Xiangxiang Chu, Zhi Tian, Yuqing Wang, Bo~Zhang, Haibing Ren, Xiaolin Wei,
  Huaxia Xia, and Chunhua Shen.
\newblock Twins: Revisiting the design of spatial attention in vision
  transformers.
\newblock {\em NeurIPS}, 34, 2021.

\bibitem{cubuk2020randaugment}
Ekin~D Cubuk, Barret Zoph, Jonathon Shlens, and Quoc~V Le.
\newblock Randaugment: Practical automated data augmentation with a reduced
  search space.
\newblock In {\em CVPRW}, pages 702--703, 2020.

\bibitem{cui2022mixformer}
Yutao Cui, Jiang Cheng, Limin Wang, and Gangshan Wu.
\newblock Mixformer: End-to-end tracking with iterative mixed attention.
\newblock {\em CVPR}, 2022.

\bibitem{dai2021coatnet}
Zihang Dai, Hanxiao Liu, Quoc~V Le, and Mingxing Tan.
\newblock Coatnet: Marrying convolution and attention for all data sizes.
\newblock {\em NeurIPS}, 34:3965--3977, 2021.

\bibitem{deng2009imagenet}
Jia Deng, Wei Dong, Richard Socher, Li-Jia Li, Kai Li, and Li~Fei-Fei.
\newblock Imagenet: A large-scale hierarchical image database.
\newblock In {\em CVPR}, pages 248--255, 2009.

\bibitem{ding2022scaling}
Xiaohan Ding, Xiangyu Zhang, Yizhuang Zhou, Jungong Han, Guiguang Ding, and
  Jian Sun.
\newblock Scaling up your kernels to 31x31: Revisiting large kernel design in
  cnns.
\newblock {\em CVPR}, 2022.

\bibitem{dong2021cswin}
Xiaoyi Dong, Jianmin Bao, Dongdong Chen, Weiming Zhang, Nenghai Yu, Lu~Yuan,
  Dong Chen, and Baining Guo.
\newblock Cswin transformer: A general vision transformer backbone with
  cross-shaped windows.
\newblock {\em CVPR}, 2022.

\bibitem{dosovitskiy2020vit}
Alexey Dosovitskiy, Lucas Beyer, Alexander Kolesnikov, Dirk Weissenborn,
  Xiaohua Zhai, Thomas Unterthiner, Mostafa Dehghani, Matthias Minderer, Georg
  Heigold, Sylvain Gelly, Jakob Uszkoreit, and Neil Houlsby.
\newblock An image is worth 16x16 words: Transformers for image recognition at
  scale.
\newblock {\em arXiv preprint arXiv:2010.11929}, 2020.

\bibitem{d2021convit}
St{\'e}phane d’Ascoli, Hugo Touvron, Matthew~L Leavitt, Ari~S Morcos, Giulio
  Biroli, and Levent Sagun.
\newblock Convit: Improving vision transformers with soft convolutional
  inductive biases.
\newblock In {\em ICML}, pages 2286--2296, 2021.

\bibitem{fan2021multiscale}
Haoqi Fan, Bo~Xiong, Karttikeya Mangalam, Yanghao Li, Zhicheng Yan, Jitendra
  Malik, and Christoph Feichtenhofer.
\newblock Multiscale vision transformers.
\newblock In {\em ICCV}, pages 6824--6835, 2021.

\bibitem{guo2022van}
Meng-Hao Guo, Cheng-Ze Lu, Zheng-Ning Liu, Ming-Ming Cheng, and Shi-Min Hu.
\newblock Visual attention network.
\newblock {\em arXiv preprint arXiv:2202.09741}, 2022.

\bibitem{han2021demystifying}
Qi~Han, Zejia Fan, Qi~Dai, Lei Sun, Ming-Ming Cheng, Jiaying Liu, and Jingdong
  Wang.
\newblock Demystifying local vision transformer: Sparse connectivity, weight
  sharing, and dynamic weight.
\newblock {\em arXiv preprint arXiv:2106.04263}, 2021.

\bibitem{he2017mask}
Kaiming He, Georgia Gkioxari, Piotr Dollar, and Ross Girshick.
\newblock Mask r-cnn.
\newblock In {\em ICCV}, 2017.

\bibitem{he2016deep}
Kaiming He, Xiangyu Zhang, Shaoqing Ren, and Jian Sun.
\newblock Deep residual learning for image recognition.
\newblock In {\em CVPR}, pages 770--778, 2016.

\bibitem{hochreiter1997lstm}
Sepp Hochreiter and J{\"u}rgen Schmidhuber.
\newblock Long short-term memory.
\newblock {\em Neural computation}, 9(8):1735--1780, 1997.

\bibitem{howard2017mobilenets}
Andrew~G Howard, Menglong Zhu, Bo~Chen, Dmitry Kalenichenko, Weijun Wang,
  Tobias Weyand, Marco Andreetto, and Hartwig Adam.
\newblock Mobilenets: Efficient convolutional neural networks for mobile vision
  applications.
\newblock {\em arXiv preprint arXiv:1704.04861}, 2017.

\bibitem{senet}
Jie Hu, Li~Shen, and Gang Sun.
\newblock Squeeze-and-excitation networks.
\newblock In {\em CVPR}, pages 7132--7141, 2018.

\bibitem{stochasticdepth}
Gao Huang, Yu~Sun, Zhuang Liu, Daniel Sedra, and Kilian~Q Weinberger.
\newblock Deep networks with stochastic depth.
\newblock In {\em ECCV}, pages 646--661, 2016.

\bibitem{huang2021fapn}
Shihua Huang, Zhichao Lu, Ran Cheng, and Cheng He.
\newblock Fapn: Feature-aligned pyramid network for dense image prediction.
\newblock In {\em ICCV}, pages 864--873, 2021.

\bibitem{jia2016dynamic}
Xu~Jia, Bert De~Brabandere, Tinne Tuytelaars, and Luc~V Gool.
\newblock Dynamic filter networks.
\newblock {\em NeurIPS}, 29, 2016.

\bibitem{jiang2021token}
Zihang Jiang, Qibin Hou, Li~Yuan, Daquan Zhou, Xiaojie Jin, Anran Wang, and
  Jiashi Feng.
\newblock Token labeling: Training a 85.5\% top-1 accuracy vision transformer
  with 56m parameters on imagenet.
\newblock {\em arXiv preprint arXiv:2104.10858}, 2021.

\bibitem{kolesnikov2020big}
Alexander Kolesnikov, Lucas Beyer, Xiaohua Zhai, Joan Puigcerver, Jessica Yung,
  Sylvain Gelly, and Neil Houlsby.
\newblock Big transfer (bit): General visual representation learning.
\newblock In {\em ECCV}, pages 491--507. Springer, 2020.

\bibitem{krizhevsky2012alex}
Alex Krizhevsky, Ilya Sutskever, and Geoffrey~E Hinton.
\newblock Imagenet classification with deep convolutional neural networks.
\newblock {\em NeurIPS}, 25:1097--1105, 2012.

\bibitem{lenet}
Yann LeCun, Bernhard Boser, John~S Denker, Donnie Henderson, Richard~E Howard,
  Wayne Hubbard, and Lawrence~D Jackel.
\newblock Backpropagation applied to handwritten zip code recognition.
\newblock {\em Neural computation}, 1(4):541--551, 1989.

\bibitem{lerman2021explaining}
Samuel Lerman, Charles Venuto, Henry Kautz, and Chenliang Xu.
\newblock Explaining local, global, and higher-order interactions in deep
  learning.
\newblock In {\em ICCV}, pages 1224--1233, 2021.

\bibitem{li2022exploring}
Yanghao Li, Hanzi Mao, Ross Girshick, and Kaiming He.
\newblock Exploring plain vision transformer backbones for object detection.
\newblock {\em arXiv preprint arXiv:2203.16527}, 2022.

\bibitem{li2022mvitv2}
Yanghao Li, Chao-Yuan Wu, Haoqi Fan, Karttikeya Mangalam, Bo~Xiong, Jitendra
  Malik, and Christoph Feichtenhofer.
\newblock Mvitv2: Improved multiscale vision transformers for classification
  and detection.
\newblock In {\em CVPR}, pages 4804--4814, 2022.

\bibitem{lin2017fpn}
Tsung-Yi Lin, Piotr Doll{\'a}r, Ross Girshick, Kaiming He, Bharath Hariharan,
  and Serge Belongie.
\newblock Feature pyramid networks for object detection.
\newblock In {\em CVPR}, pages 2117--2125, 2017.

\bibitem{lin2017feature}
Tsung-Yi Lin, Piotr Doll{\'a}r, Ross Girshick, Kaiming He, Bharath Hariharan,
  and Serge Belongie.
\newblock Feature pyramid networks for object detection.
\newblock In {\em CVPR}, pages 2117--2125, 2017.

\bibitem{lin2014coco}
Tsung-Yi Lin, Michael Maire, Serge Belongie, James Hays, Pietro Perona, Deva
  Ramanan, Piotr Doll{\'a}r, and C~Lawrence Zitnick.
\newblock Microsoft coco: Common objects in context.
\newblock In {\em ECCV}, pages 740--755. Springer, 2014.

\bibitem{gmlp}
Hanxiao Liu, Zihang Dai, David So, and Quoc~V Le.
\newblock Pay attention to mlps.
\newblock {\em NeurIPS}, 34:9204--9215, 2021.

\bibitem{liu2022dynamic}
Kai Liu, Tianyi Wu, Cong Liu, and Guodong Guo.
\newblock Dynamic group transformer: A general vision transformer backbone with
  dynamic group attention.
\newblock {\em IJCAI}, 2022.

\bibitem{liu2021swinv2}
Ze~Liu, Han Hu, Yutong Lin, Zhuliang Yao, Zhenda Xie, Yixuan Wei, Jia Ning, Yue
  Cao, Zheng Zhang, Li~Dong, et~al.
\newblock Swin transformer v2: Scaling up capacity and resolution.
\newblock {\em arXiv preprint arXiv:2111.09883}, 2021.

\bibitem{liu2021swin}
Ze~Liu, Yutong Lin, Yue Cao, Han Hu, Yixuan Wei, Zheng Zhang, Stephen Lin, and
  Baining Guo.
\newblock Swin transformer: Hierarchical vision transformer using shifted
  windows.
\newblock {\em arXiv preprint arXiv:2103.14030}, 2021.

\bibitem{liu2022convnet}
Zhuang Liu, Hanzi Mao, Chao-Yuan Wu, Christoph Feichtenhofer, Trevor Darrell,
  and Saining Xie.
\newblock A convnet for the 2020s.
\newblock {\em CVPR}, 2022.

\bibitem{adamw}
Ilya Loshchilov and Frank Hutter.
\newblock Decoupled weight decay regularization.
\newblock {\em arXiv preprint arXiv:1711.05101}, 2017.

\bibitem{acmix}
Xuran Pan, Chunjiang Ge, Rui Lu, Shiji Song, Guanfu Chen, Zeyi Huang, and Gao
  Huang.
\newblock On the integration of self-attention and convolution.
\newblock {\em arXiv preprint arXiv:2111.14556}, 2021.

\bibitem{rao2021global}
Yongming Rao, Wenliang Zhao, Zheng Zhu, Jiwen Lu, and Jie Zhou.
\newblock Global filter networks for image classification.
\newblock In {\em NeurIPS}, 2021.

\bibitem{ridnik2021imagenet}
Tal Ridnik, Emanuel Ben-Baruch, Asaf Noy, and Lihi Zelnik-Manor.
\newblock Imagenet-21k pretraining for the masses.
\newblock {\em arXiv:2104.10972}, 2021.

\bibitem{riquelme2021scaling}
Carlos Riquelme, Joan Puigcerver, Basil Mustafa, Maxim Neumann, Rodolphe
  Jenatton, Andr{\'e} Susano~Pinto, Daniel Keysers, and Neil Houlsby.
\newblock Scaling vision with sparse mixture of experts.
\newblock {\em NeurIPS}, 34, 2021.

\bibitem{simonyan2014very}
Karen Simonyan and Andrew Zisserman.
\newblock Very deep convolutional networks for large-scale image recognition.
\newblock {\em arXiv preprint arXiv:1409.1556}, 2014.

\bibitem{szegedy2015going}
Christian Szegedy, Wei Liu, Yangqing Jia, Pierre Sermanet, Scott Reed, Dragomir
  Anguelov, Dumitru Erhan, Vincent Vanhoucke, and Andrew Rabinovich.
\newblock Going deeper with convolutions.
\newblock In {\em CVPR}, pages 1--9, 2015.

\bibitem{tan2019efficientnet}
Mingxing Tan and Quoc Le.
\newblock Efficientnet: Rethinking model scaling for convolutional neural
  networks.
\newblock In {\em ICML}, pages 6105--6114, 2019.

\bibitem{touvron2020deit}
Hugo Touvron, Matthieu Cord, Matthijs Douze, Francisco Massa, Alexandre
  Sablayrolles, and Herv{\'e} J{\'e}gou.
\newblock Training data-efficient image transformers \& distillation through
  attention.
\newblock {\em arXiv preprint arXiv:2012.12877}, 2020.

\bibitem{touvron2021going}
Hugo Touvron, Matthieu Cord, Alexandre Sablayrolles, Gabriel Synnaeve, and
  Herv{\'e} J{\'e}gou.
\newblock Going deeper with image transformers.
\newblock {\em arXiv preprint arXiv:2103.17239}, 2021.

\bibitem{tu2022maxim}
Zhengzhong Tu, Hossein Talebi, Han Zhang, Feng Yang, Peyman Milanfar, Alan
  Bovik, and Yinxiao Li.
\newblock Maxim: Multi-axis mlp for image processing.
\newblock In {\em CVPR}, pages 5769--5780, 2022.

\bibitem{tu2022maxvit}
Zhengzhong Tu, Hossein Talebi, Han Zhang, Feng Yang, Peyman Milanfar, Alan
  Bovik, and Yinxiao Li.
\newblock Maxvit: Multi-axis vision transformer.
\newblock {\em ECCV}, 2022.

\bibitem{Vaswani2017transformer}
Ashish Vaswani, Noam Shazeer, Niki Parmar, Jakob Uszkoreit, Llion Jones,
  Aidan~N Gomez, Lukasz Kaiser, and Illia Polosukhin.
\newblock Attention is all you need.
\newblock In {\em NeurIPS}, pages 5998--6008, 2017.

\bibitem{wang2020linformer}
Sinong Wang, Belinda~Z Li, Madian Khabsa, Han Fang, and Hao Ma.
\newblock Linformer: Self-attention with linear complexity.
\newblock {\em arXiv preprint arXiv:2006.04768}, 2020.

\bibitem{wang2021pyramid}
Wenhai Wang, Enze Xie, Xiang Li, Deng-Ping Fan, Kaitao Song, Ding Liang, Tong
  Lu, Ping Luo, and Ling Shao.
\newblock Pyramid vision transformer: A versatile backbone for dense prediction
  without convolutions.
\newblock In {\em ICCV}, 2021.

\bibitem{DBLP:conf/iccv/SORT}
Yan Wang, Lingxi Xie, Chenxi Liu, Siyuan Qiao, Ya~Zhang, Wenjun Zhang, Qi~Tian,
  and Alan~L. Yuille.
\newblock {SORT:} second-order response transform for visual recognition.
\newblock In {\em {IEEE} International Conference on Computer Vision, {ICCV}
  2017, Venice, Italy, October 22-29, 2017}, pages 1368--1377. {IEEE} Computer
  Society, 2017.

\bibitem{wu2021cvt}
Haiping Wu, Bin Xiao, Noel Codella, Mengchen Liu, Xiyang Dai, Lu~Yuan, and Lei
  Zhang.
\newblock Cvt: Introducing convolutions to vision transformers.
\newblock {\em arXiv preprint arXiv:2103.15808}, 2021.

\bibitem{wu2022pale}
Sitong Wu, Tianyi Wu, Haoru Tan, and Guodong Guo.
\newblock Pale transformer: A general vision transformer backbone with
  pale-shaped attention.
\newblock In {\em AAAI}, volume~36, pages 2731--2739, 2022.

\bibitem{xiao2018unified}
Tete Xiao, Yingcheng Liu, Bolei Zhou, Yuning Jiang, and Jian Sun.
\newblock Unified perceptual parsing for scene understanding.
\newblock In {\em ECCV}, pages 418--434, 2018.

\bibitem{xiao2021early}
Tete Xiao, Mannat Singh, Eric Mintun, Trevor Darrell, Piotr Doll{\'a}r, and
  Ross Girshick.
\newblock Early convolutions help transformers see better.
\newblock {\em NeurIPS}, 34:30392--30400, 2021.

\bibitem{yan2022multiview}
Shen Yan, Xuehan Xiong, Anurag Arnab, Zhichao Lu, Mi~Zhang, Chen Sun, and
  Cordelia Schmid.
\newblock Multiview transformers for video recognition.
\newblock {\em arXiv preprint arXiv:2201.04288}, 2022.

\bibitem{yang2022focal}
Jianwei Yang, Chunyuan Li, and Jianfeng Gao.
\newblock Focal modulation networks.
\newblock {\em arXiv preprint arXiv:2203.11926}, 2022.

\bibitem{yang2021focal}
Jianwei Yang, Chunyuan Li, Pengchuan Zhang, Xiyang Dai, Bin Xiao, Lu~Yuan, and
  Jianfeng Gao.
\newblock Focal attention for long-range interactions in vision transformers.
\newblock {\em NeurIPS}, 34, 2021.

\bibitem{yu2021metaformer}
Weihao Yu, Mi~Luo, Pan Zhou, Chenyang Si, Yichen Zhou, Xinchao Wang, Jiashi
  Feng, and Shuicheng Yan.
\newblock Metaformer is actually what you need for vision.
\newblock {\em arXiv preprint arXiv:2111.11418}, 2021.

\bibitem{yuan2021t2t}
Li~Yuan, Yunpeng Chen, Tao Wang, Weihao Yu, Yujun Shi, Zihang Jiang, Francis~EH
  Tay, Jiashi Feng, and Shuicheng Yan.
\newblock Tokens-to-token vit: Training vision transformers from scratch on
  imagenet.
\newblock {\em arXiv:2101.11986}, 2021.

\bibitem{zagoruyko2016wide}
Sergey Zagoruyko and Nikos Komodakis.
\newblock Wide residual networks.
\newblock {\em arXiv preprint arXiv:1605.07146}, 2016.

\bibitem{zhang2022dino}
Hao Zhang, Feng Li, Shilong Liu, Lei Zhang, Hang Su, Jun Zhu, Lionel~M Ni, and
  Heung-Yeung Shum.
\newblock Dino: Detr with improved denoising anchor boxes for end-to-end object
  detection.
\newblock {\em arXiv preprint arXiv:2203.03605}, 2022.

\bibitem{zhou2017scene}
Bolei Zhou, Hang Zhao, Xavier Puig, Sanja Fidler, Adela Barriuso, and Antonio
  Torralba.
\newblock Scene parsing through ade20k dataset.
\newblock In {\em CVPR}, pages 633--641, 2017.

\end{thebibliography}
}

\newpage

\begin{appendix}
\section{FLOPs of \gnconv{}} \label{appendix:flops}
We will divide the computation of our \gnconv{} into 3 parts, and calculate the FLOPs for each part.
\begin{itemize}
    \item \textbf{Projection layers.} The FLOPs of two projection layers $\phi_{\rm in}$ and $\phi_{\rm out}$ can be easily derived as:
    \begin{equation}
            \flops(\phi_{\rm in}) = 2HWC^2, \quad \flops(\phi_{\rm out}) = HWC^2
    \end{equation}
    \item \textbf{Depth-wise convolution.} We first consider the standard depth-wise convolution (DWConv) with kernel size $K$. The DWConv is performed for all $\{\mathbf{q}_k\}_{k=1}^{n-1}$, where $\mathbf{q}_k\in \mathbb{R}^{HW\times C_k}$ and $C_k=\frac{C}{2^{n-k-1}}$. Therefore, the FLOPs for DWConv are
    \begin{equation}
    \begin{split}
        \flops(\text{DWConv}) &= HWK^2\sum_{k=0}^{n-1}\frac{C}{2^{n-k-1}}=2HWCK^2\left(1 - \frac{1}{2^n}\right).
    \end{split}
    \end{equation}
    \item \textbf{Recursive Gating.} We consider both the flops of the projection layer $g_k$ and the element-wise multiplication.
    \begin{equation}
    \begin{split}
        \flops(\text{RecursiveGating}) &=HWC_0 + \sum_{k=1}^{n-1}(HWC_{k-1}C_k + HWC_k) \\
        &=HWC\left[\frac{2}{3}C\left(1 - \frac{1}{4^{n-1}}\right) + 2 - \frac{1}{2^{n-1}}\right].
    \end{split}
    \end{equation}
\end{itemize}
Therefore, the total FLOPs are:
\begin{equation}
    \flops(\gnconv{}) = HWC\left[2K^2\left(1 - \frac{1}{2^n}\right) + \left(\frac{11}{3} - \frac{2}{3\times 4^{n-1}}\right)C + 2 - \frac{1}{2^{n-1}}\right].
\end{equation}

\section{Spatial Interactions in Vision Models. } \label{appendix:highorder}

We review some representative vision model designs from the perspective of spatial interactions, as shown in Figure 1. Specifically, we are interested in the interactions between a feature $\mathbf{x}_i$ and its neighbor feature $\mathbf{x}_j, j\in\Omega_i$. Inspired by the interaction effect (IE)~\cite{lerman2021explaining,ai2003interaction}, we consider that a binary function $F(\mathbf{x}_i,\mathbf{x}_j)$ which directly operates on $\mathbf{x}_i,\mathbf{x}_j$ introduces an effective interaction between $\mathbf{x}_i$ $\mathbf{x}_j$, if
\begin{equation}
    \IE(F)=\frac{\partial F}{\partial \mathbf{x}_i \partial \mathbf{x}_j} \neq \mathbf{0}. \label{equ:ie}
\end{equation}
We now analyze the cases in Figure 1 of our main paper using the above rule. \textbf{(a): Convolution.} The output $F_i=\sum_{j\in\Omega}w_{i\to j}\mathbf{x}_j$, which leads to $\IE(F)=\mathbf{0}$. Therefore, standard convolution introduce no interaction between $\mathbf{x}_i$ and $\mathbf{x}_j$ and we call it a \textit{0-order interaction}. \textbf{(b): SE Block/Gated Convolution.} In this case, we have $F_i=\sum_{j\in\Omega}w_{i\to j}\mathbf{x}_j s_i(\mathbf{x})$, where $s_i(\mathbf{x})=\frac{1}{HW}\sum_{l=1}^{HW}x_{l}$ for the SE block and $s_i(\mathbf{x})=\mathbf{x}_i$ for the gated convolution. It is easy to show $\IE(F)\neq \mathbf{0}$ because $\frac{\partial s_i}{\partial \mathbf{x}_i}\neq \mathbf{0}$. Hence, these two operations both introduce \textit{1-order interaction}. \textbf{(c): Self-attention (SA).} We first denote the projected query/key/value features as $\mathbf{q},\mathbf{k},\mathbf{v}$. The SA first perform an 1-order interaction by computing the attention with dot-product: $\mathbf{a}_i = \mathbf{q}_i^{\top}[\mathbf{k}_1, \ldots, \mathbf{k}_{HW}]/\sqrt{C}$. We then view $\mathbf{a}_i$ as the feature at location $i$ in the following computation. The normalized $\hat{\mathbf{a}}_i$ is then obtained by Softmax, which do not contribute to the order since it can be viewed as an implicit interaction that does not explicitly introduce $\mathbf{x}_j$ to the computation. The second interaction is performed by $\mathbf{x}_i = \sum_{j\in\Omega}\hat{\mathbf{a}}_i\mathbf{v}_j$. To sum up, the SA is a \textit{2-order interaction}. \textbf{(d)}: \gnconv{}. According to Section 3.1, we have already known that \gnconv{} can achieve $n$-order interaction with bounded computational cost.

From the above discussion, we reveal a key difference between ViTs and previous architectures from a new view, \ie, ViTs have higher-order spatial interactions in each basic block. Then it begs the question that whether we can achieve better accuracy-complexity trade-offs viz interactions with more than 2 orders. Our proposed \gnconv{} exactly targets this question for the first time. First, we can realize arbitrary $n$-order interaction as long as $1\le n \le 1 + \log_2 C$ easily. Second, unlike the quadratic complexity of self-attention, the computational cost of \gnconv{} has an upper bound \wrt the order $n$.

 In our implementation of \gnconv{}, the higher-order spatial interactions are based on the gating mechanism, which has also been investigated in LSTM~\cite{hochreiter1997lstm} and some vision modules~\cite{DBLP:conf/iccv/SORT,tu2022maxim,gmlp}. However, these previous methods can only achieve up to 2-order interactions, and did not fully reveal the potential of higher-order interactions. On the contrary, our \gnconv{} is more extendable to achieve arbitrary higher-order spatial interactions under a controllable computational budget.

\section{Implementation Details} \label{appendix:train_details}

\subsection{Architecture Details. }

To better verify the effectiveness of our new designs, we introduce minimal changes in the overall architecture of Swin Transformers~\cite{liu2021swin}. Specifically, we make two changes to the overall architecture of Swin Transformers~\cite{liu2021swin}: 1) We add one block in stage 2 to make the overall computation and parameters close to previous models; 2) We use the LayerScale~\cite{touvron2021going} techniques to make our models more stable during training following the practice of ConvNeXt~\cite{liu2022convnet}. Note that the two changes have been applied to the baseline model considered in our ablation study to clearly show the effects of our designs. The detailed architectures of ConvNeXt~\cite{liu2022convnet}, Swin Transformers~\cite{liu2021swin} and HorNet are summarized in Table~\ref{table:arch}.

\begin{table}[t]
  \centering
  \caption{\small {The detailed architectures of ConvNeXt~\cite{liu2022convnet}, Swin Transformers~\cite{liu2021swin}, and HorNet.}}\label{table:arch}
  \adjustbox{width=\linewidth}{
    \begin{tabular}{C{40pt}|C{50pt}|C{120pt}|C{120pt}|C{120pt}}
    \toprule
          &  \multirow{2}[0]{*}{Output Size} & \textbf{ConvNeXt-S/B/L} & \textbf{Swin-S/B/L} & \textbf{HorNet-T/S/B/L} \\
          & & $C$=96/128/192 &  $C$=96/128/192 &  $C$=64/96/128/192 \\
           \midrule
   \multirow{1}[0]{*}{Stem} & 56$\times$56 & Conv$_{\text{4}\times\text{4}}$, $C$, stride 4 & Conv$_{\text{4}\times\text{4}}$, $C$, stride 4  & Conv$_{\text{4}\times\text{4}}$, $C$, stride 4  \\
   \midrule
    \multirow{3}[0]{*}{Stage1} &  
    \multirow{3}[0]{*}{56$\times$56} & 
    \multirow{3}[0]{*}{$\begin{bmatrix}\text{DWConv}_{\text{7}\times\text{7}}, C\\\text{MLP}, 4C, C \end{bmatrix}$ $\times$ 3} &
     \multirow{3}[0]{*}{$\begin{matrix}\begin{bmatrix}\text{MSA}_{\text{7}\times\text{7}}^{H=C/32}, C  \end{bmatrix}\\ \begin{bmatrix} \text{MLP}, 4C, C \end{bmatrix} \end{matrix}$ $\times$ 2} &
      \multirow{3}[0]{*}{$\begin{matrix} \begin{bmatrix} \text{\ensuremath{\textit{g}^\textit{2}\text{Conv}}}_{\text{7}\times\text{7}/{\rm GF}}, C \end{bmatrix}\\ \begin{bmatrix} \text{MLP}, 4C, C \end{bmatrix} \end{matrix}$ $\times$ 2} \\
    & & & & \\ 
     & & & & \\ 
    \midrule
    \multirow{3}[0]{*}{Stage2} &  
    \multirow{3}[0]{*}{28$\times$28} & 
    \multirow{3}[0]{*}{$\begin{bmatrix}\text{DWConv}_{\text{7}\times\text{7}}, 2C\\\text{MLP}, 8C, 2C \end{bmatrix}$ $\times$ 3} &
     \multirow{3}[0]{*}{$\begin{matrix}\begin{bmatrix}\text{MSA}_{\text{7}\times\text{7}}^{H=C/32}, 2C  \end{bmatrix}\\ \begin{bmatrix} \text{MLP}, 8C, 2C \end{bmatrix} \end{matrix}$ $\times$ 2} &
      \multirow{3}[0]{*}{$\begin{matrix} \begin{bmatrix} \text{\ensuremath{\textit{g}^\textit{3}\text{Conv}}}_{\text{7}\times\text{7}/{\rm GF}}, 2C \end{bmatrix}\\ \begin{bmatrix} \text{MLP}, 8C, 2C \end{bmatrix} \end{matrix}$ $\times$ 3} \\
    & & & & \\ 
     & & & & \\ 
    \midrule
    \multirow{3}[0]{*}{Stage3} &  
    \multirow{3}[0]{*}{14$\times$14} & 
    \multirow{3}[0]{*}{$\begin{bmatrix}\text{DWConv}_{\text{7}\times\text{7}}, 4C\\\text{MLP}, 16C, 4C \end{bmatrix}$ $\times$ 27} &
     \multirow{3}[0]{*}{$\begin{matrix}\begin{bmatrix}\text{MSA}_{\text{7}\times\text{7}}^{H=C/32}, 4C  \end{bmatrix}\\ \begin{bmatrix} \text{MLP}, 16C, 4C \end{bmatrix} \end{matrix}$ $\times$ 18} &
      \multirow{3}[0]{*}{$\begin{matrix} \begin{bmatrix} \text{\ensuremath{\textit{g}^\textit{4}\text{Conv}}}_{\text{7}\times\text{7}/{\rm GF}}, 4C \end{bmatrix}\\ \begin{bmatrix} \text{MLP}, 16C, 4C \end{bmatrix} \end{matrix}$ $\times$ 18} \\
    & & & & \\ 
     & & & & \\ 
    \midrule
    \multirow{3}[0]{*}{Stage4} &  
    \multirow{3}[0]{*}{7$\times$7} & 
    \multirow{3}[0]{*}{$\begin{bmatrix}\text{DWConv}_{\text{7}\times\text{7}}, 8C\\\text{MLP}, 32C, 8C \end{bmatrix}$ $\times$ 3} &
     \multirow{3}[0]{*}{$\begin{matrix}\begin{bmatrix}\text{MSA}_{\text{7}\times\text{7}}^{H=C/32}, 8C  \end{bmatrix}\\ \begin{bmatrix} \text{MLP}, 32C, 8C \end{bmatrix} \end{matrix}$ $\times$ 2} &
      \multirow{3}[0]{*}{$\begin{matrix} \begin{bmatrix} \text{\ensuremath{\textit{g}^\textit{5}\text{Conv}}}_{\text{7}\times\text{7}/{\rm GF}}, 8C \end{bmatrix}\\ \begin{bmatrix} \text{MLP}, 32C, 8C \end{bmatrix} \end{matrix}$ $\times$ 2} \\
    & & & & \\ 
     & & & & \\ \midrule
     Classifier & & \multicolumn{3}{c}{Global Average Pooling, Linear}\\
     
     \bottomrule
    
    \end{tabular}%
    }
\end{table}%

\subsection{Experimental Settings for Image Classification. }

\paragrapha{ImageNet-1K training. } ImageNet-1K~\cite{deng2009imagenet} is a widely used large-scale benchmark for image classification, which contains around 1.2 million images from 1,000 categories. Following common practice~\cite{he2016deep,liu2022convnet}, we train our models on the training set of ImageNet and report the single-crop top-1 accuracy on 50,000 validation images. To fairly compare with our baseline methods (\ie, Swin Transformers~\cite{liu2021swin} and ConvNeXt~\cite{liu2022convnet}), we follow the most training details of ConvNeXt and make several small modifications to make the training configurations suitable for our models. For HorNet with 7$\times$7 convolutions, we find that applying gradient clipping with a maximal norm of 5 will significantly stabilize the training process, which may be due to the large gradients brought by the high-order structures in our models. For HorNet with global filters, we use stronger regularization strategies since we find that larger kernels will improve the model capacity but may also cause more severe overfitting. Specifically, we set the gradient norm to 1 and use more aggressive RandAug~\cite{cubuk2020randaugment} data augmentation strategies (\ie, we adjust the magnitudes for tiny, small and base models to 9, 12 and 15, respectively). We set the stochastic depth coefficient of HorNet-T/S/B models to 0.2, 0.4 and 0.5.  The other details are identical to ConvNeXt~\cite{liu2022convnet}. Our models are trained using 32 NVIDIA A100 GPUs with a global batch size of 4096.

\paragrapha{ImageNet-22K training. } ImageNet-22K~\cite{deng2009imagenet} is a larger dataset that contains $>$21k classes and around 14M images. We use the subset suggested by~\cite{ridnik2021imagenet} since the new \texttt{winter 2021 release
} is the accessible version now. We also follow the~\cite{ridnik2021imagenet} to remove categories with few images, resulting in roughly half fewer categories and only
13\% fewer images compared to the original dataset.  We follow previous practice~\cite{liu2021swin, liu2022convnet} to train our models for 90 epochs and use a similar data augmentation strategy as ImageNet-1K experiments. We set the stochastic depth coefficient~\cite{stochasticdepth} to 0.2. We also set the maximal gradient norm to 5 and 1 for our large models with standard 7$\times$7 convolutions and global filters respectively. We also adjust the weight decay to 0.1. The other details are identical to ConvNeXt~\cite{liu2022convnet}. We also fine-tune our best model HorNet-L$_{\rm GF}$ on 384$\times$384 images on ImageNet-22K for 10 epochs compete with state-of-the-art models on downstream tasks. The model is only used in the experiments in Appendix~\ref{appendix:sota}.

\paragrapha{ImageNet-1K fine-tuning. } We fine-tune the models pre-trained on ImageNet-22K or at the 224$\times$224 resolution to ImageNet-1K or/and 384$\times$384 resolution for 30 epochs with a batch size of 512 and a cosine learning rate schedule with an initial learning rate of $5e^{-5}$. We set the weight decay to $1e^{-6}$ and disable MixUp and CutMix following~\cite{liu2022convnet}. We initialize the ImageNet-1K classifier with the corresponding classifier weights for ImageNet-22K classes to further stabilize the training process.

\subsection{Experimental Settings for Downstream Tasks.}
\paragrapha{Object detection and instance segmentation on COCO.} We adopt the widely used Cascade Mask R-CNN~\cite{cai2018cascade} framework to perform object detection and instance segmentation on COCO, following Swin~\cite{liu2021swin} and ConvNeXt~\cite{liu2022convnet}. Our backbones are pre-trained on ImageNet-1K for the HorNet-T/S/B and ImageNet-22K for the HorNet-L. We use the 3$\times$ schedule where we train all of our model for 36 epochs with AdamW~\cite{adamw} optimizer and a global batch size of 16. We set the learning rate of as \{2e-4, 2e-4, 2e-4, 1e-4\} and the stochastic depth rate as \{0.4, 0.6, 0.7, 0.7\}for HorNet-T/S/B/L. We set the weight decay as 0.05 for all the models.

\paragrapha{Semantic Segmentation on ADE20K.} We use the UperNet 160K~\cite{xiao2018unified} framework for semantic segmentation on ADE20K. We use a global batch size of 16 and train all the models for 160 iterations with the AdamW~\cite{adamw} optimizer. We use $512\times 512$ image for ImageNet-1K pre-trained HorNet-T/S/B and $640\times 640$ image for ImagNet-22K pre-trained HorNet-L. We set the learning rate as 1e-4 and the weight decay as 0.05 for all the models. We report the mIoU of both single-scale and multi-scale testing on the validation set.

\section{More Analysis}

\begin{table}[t]
 \centering 
 \caption{\textbf{Throughput analysis. } We provide the detailed throughput statistics of our models and several baseline methods. The throughput is measured with a single NVIDIA RTX 3090 GPU with a batch size of 128. }
 \label{tab:speed}
  \begin{tabular}{L{80pt}L{50pt}L{50pt}L{50pt}}
  \toprule 
  \multirow{2}[2]{*}{Model} & FLOPs & Throughput & Top-1 Acc.\\ 
  & (G) & (img/s) & (\%) \\
  \midrule
 ConvNeXt-T~\cite{liu2022convnet} & 4.5 &   1010.3  & 82.1 \\
 Swin-T~\cite{liu2021swin} & 4.5 &   832.2  & 81.3  \\
 MViTv2-T~\cite{li2022mvitv2} & 4.7 &   728.4  & 82.3 \\
 \rowcolor{Gray}HorNet-T$_{7\times 7}$ &  4.0 &   845.7 & 82.8 \\
 \midrule
 ConvNeXt-S~\cite{liu2022convnet} & 8.7 &  621.5  & 83.1 \\
 Swin-S~\cite{liu2021swin} &  8.7 &   520.7  & 83.0 \\
 MViTv2-S~\cite{li2022mvitv2} & 7.0 &  531.5 & 83.6 \\
 \rowcolor{Gray} HorNet-S$_{7\times 7}$ &  8.8 &   525.8 & 83.8 \\
 \midrule
ConvNeXt-B~\cite{liu2022convnet} & 15.4 &  440.8  & 83.8 \\
Swin-B~\cite{liu2021swin} & 15.4 &   364.8  & 83.5 \\
MViTv2-B~\cite{li2022mvitv2} & 10.2 &  369.1  & 84.4 \\
\rowcolor{Gray} HorNet-B$_{7\times 7}$ & 15.6 &   410.0 & 84.2 \\ 
  \bottomrule
  \end{tabular}%
\end{table}%

\begin{table}[t]
 \centering 
 \caption{Effects of $\alpha$.}
 \label{tab:alpha}
  \begin{tabular}{L{130pt}L{35pt}L{35pt}L{35pt}L{35pt}L{35pt}}
  \toprule 
  $\alpha$ & 1 & 2 & 3 & 5 & 10 \\
  \midrule
  ImageNet Top-1 Acc. (\%) & 82.71 & 82.76 & \textbf{82.81} & 82.74 & 82.69 \\
  \bottomrule
  \end{tabular}%

\end{table}%

\begin{table}[t]
 \centering 
 \caption{Effects of activation functions in gated convolutions.}
 \label{tab:activation}
  \begin{tabular}{L{130pt}L{45pt}L{45pt}L{45pt}L{45pt}}
  \toprule 
  & None & \texttt{Sigmoid} & \texttt{Tanh} & \texttt{GELU} \\
  \midrule
  ImageNet Top-1 Acc. (\%) & 82.7 & 82.3$_{\cb{\text{(-0.4)}}}$ & 82.6$_{\cb{\text{(-0.1)}}}$ & 82.2$_{\cb{\text{(-0.5)}}}$ \\
  \bottomrule
  \end{tabular}%

\end{table}%

\paragrapha{Comparisons with state-of-the-art methods on ImageNet. } Our experiments are designed to clearly verify the superior of our design over previous basic operations like plain convolution and self-attention. Therefore, we choose to  follow the basic architecture and the training configuration of widely used architectures Swin Transformers and ConvNeXt   Therefore, there is still substantial room to further improve the performance on ImageNet-1K. We notice that some recent work like Pale Transformers~\cite{wu2022pale} and Dynamic Group Transformer~\cite{liu2022dynamic} with hybrid architectures or more careful designs achieve better performance than HorNet on ImageNet-1K. We think many techniques that have been used in previous work can be useful to further imporve our models, including further optimized overall architectures (\eg, optimized depth/width for each stage), better patch embedding strategies (\eg, overlapping convolutional layers for input embedding and downsampling), more efficient ways to compute adaptive weights (\eg, using downsmapled features to produce attention weights like~\cite{li2022mvitv2}), and more advanced training methods and hybrid architectures (\eg, combining $\textit{g}^\textit{n}\text{Conv}$ with self-attention and plain convolutions).

\paragrapha{Throughput analysis. } We provide the detailed throughput statistics of our models and several baseline methods in Table~\ref{tab:speed}. Apart from ConvNeXt and Swin Transformers, we also compare our model with recent MViTv2~\cite{li2022mvitv2} models.  The multiple small matrix multiplications introduced by \gnconv{} will affect the speed of our models on GPU.  We  observed that our method is slower than ConvNeXt by 7\%~15\% with similar FLOPs. Meanwhile, thanks to the highly efficient depth-wise convolutions implementation of CuDNN, we also see that our models achieve similar or slightly faster speed than typical vision Transformers with similar FLOPs. Notably, as shown in Figure 3(c), the higher classification accuracy helps our models achieve better speed-accuracy trade-offs than ConvNeXt and Swin Transformers. Therefore, we believe the speed of our method is still competitive with these recent models. 

\paragrapha{Effects of $\alpha$. } We find that re-scaling the output of gated convolution will avoid the large values produced by the recursive process and stabilize the training process. We analyze the the effects of $\alpha$ on our ImageNet experiments based on HorNet-B$_{7\times 7}$. The results are summarized in Table~\ref{tab:alpha}. We see $\alpha=3$ leads the best performance. Therefore, we set $\alpha$ to 3 in all our models.

\paragrapha{Effects of activation functions in gated convolutions. } The gated convolutions used in our models can be viewed as a type of channel attention that uses different attention weights in different locations and generates weights based on spatial interactions. Previous channel attention methods like SE-Net~\cite{senet} usually add a \texttt{sigmoid} function to the attention weights to generate bounded attention. Therefore, we investigate several possible activation functions in our models. The results are presented in Table~\ref{tab:activation}. We see the version described in our paper (\ie, no activation function) achieves the best performance. The result also suggests that \gnconv{} exhibits a different behavior from conventional channel attention methods. Since the gating weights are critical components in our models, activation functions that can cause significant information losses like \texttt{GELU} and \texttt{sigmoid} will severely hurt the performance.

\end{appendix}

\end{document}